\documentclass[sigconf]{acmart}

\usepackage{booktabs} 
\usepackage{enumitem}
\usepackage{subcaption}
\usepackage{hyperref}
\usepackage[ruled,noend]{algorithm2e} 

\copyrightyear{2019}
\acmYear{2019} 
\setcopyright{iw3c2w3}
\acmConference[WWW '19]{Proceedings of the 2019 World Wide Web Conference}{May 13--17, 2019}{San Francisco, CA, USA}
\acmBooktitle{Proceedings of the 2019 World Wide Web Conference (WWW'19), May 13--17, 2019, San Francisco, CA, USA}
\acmPrice{}
\acmDOI{10.1145/3308558.3313394}
\acmISBN{978-1-4503-6674-8/19/05}

\usepackage{balance}

\begin{document}

\title{CBHE: Corner-based Building Height Estimation for Complex Street Scene Images}

\author{Yunxiang Zhao}
\orcid{1234-5678-9012}
\affiliation{%
	\institution{The University of Melbourne \& NUDT}
	\postcode{3053}
}
\email{yunxiangz@student.unimelb.edu.au}

\author{Jianzhong Qi}
\orcid{1234-5678-9012}
\affiliation{%
	\institution{The University of Melbourne}
	\postcode{3053}
}
\email{jianzhong.qi@unimelb.edu.au}

\author{Rui Zhang}
\authornote{Corresponding author.}
\orcid{1234-5678-9012}
\affiliation{%
	\institution{The University of Melbourne}
	\postcode{3053}
}
\email{rui.zhang@unimelb.edu.au}

\renewcommand{\shortauthors}{Y. Zhao et al.}

\begin{abstract}
	Building height estimation is important in many applications such as 3D city reconstruction, urban planning, and navigation. Recently, a new building height estimation method using street scene images and 2D maps was proposed. This method is more scalable than traditional methods that use high-resolution optical data, LiDAR data, or RADAR data which are expensive to obtain. The method needs to detect building rooflines and then compute building height via the pinhole camera model. We observe that this method has limitations in handling complex street scene images in which buildings overlap with each other and the rooflines are difficult to locate. We propose CBHE, a building height estimation algorithm considering both building corners and rooflines. CBHE first obtains building corner and roofline candidates in street scene images based on building footprints from 2D maps and the camera parameters. Then, we use a deep neural network named BuildingNet to classify and filter corner and roofline candidates. Based on the valid corners and rooflines from BuildingNet, CBHE computes building height via the pinhole camera model. Experimental results show that the proposed BuildingNet yields a higher accuracy on building corner and roofline candidate filtering compared with the state-of-the-art open set classifiers. Meanwhile, CBHE outperforms the baseline algorithm by over 10\% in building height estimation accuracy.
\end{abstract}

%
%



\maketitle


\section{Introduction}
\label{sec:introduction}
Building height plays an essential role in many applications, such as 3D city reconstruction~\cite{pan2015inferring, Armagan2017}, urban planning~\cite{ng2009policies}, navigation~\cite{Grabler2008,rousell17}, and geographic knowledge bases~\cite{gkb17}. For example, in navigation, knowing the height of buildings helps identify those standing out in a city block, which can then be used to facilitate navigation by generating instructions such as "turn left before an 18 meters high (five-story) building". 

\begin{figure}[tp!]
	\centering
	\setlength{\abovecaptionskip}{1pt}   
	\setlength{\belowcaptionskip}{-2pt}   
	\includegraphics[width=1\linewidth]{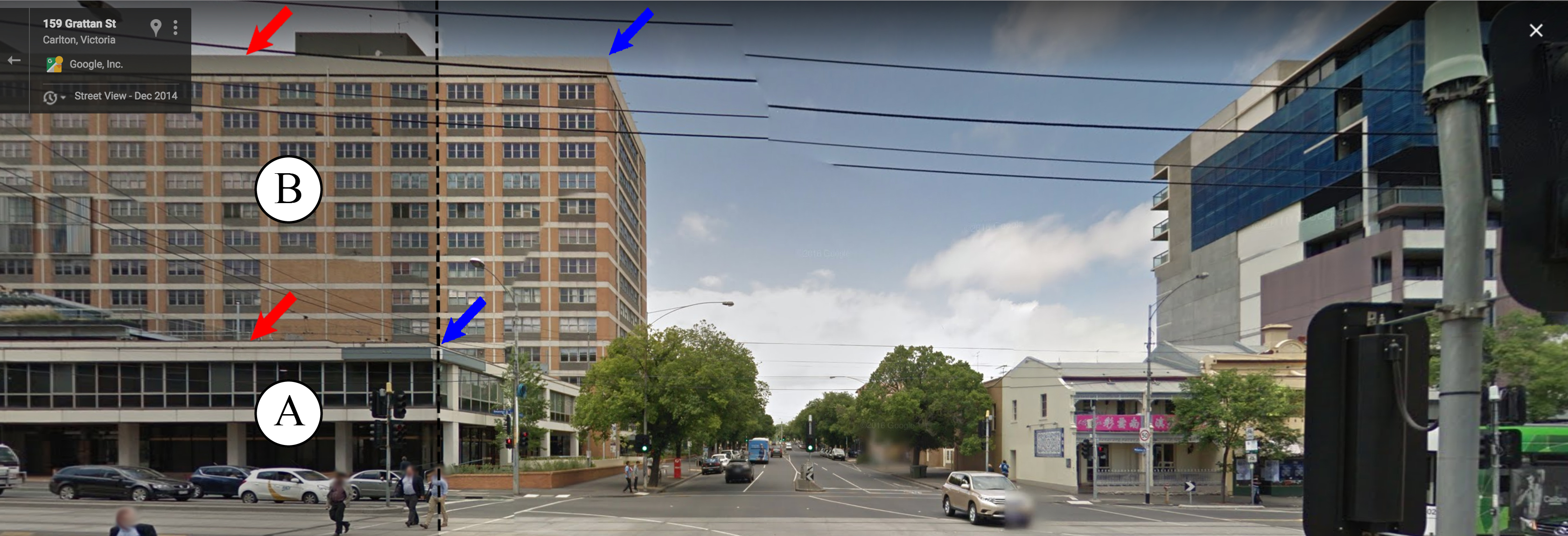}
	\caption{Building A's rooflines are hard to detect due to the overlapping with building B, while building A's corners can help figure out the true rooflines. Red arrows: building rooflines; Blue arrows: building corners.}
	\label{fig:illustration}
\end{figure}

Previous studies for building height estimation are mainly based on high-resolution optical data~\cite{izadi2012,zeng2014}, synthetic aperture radar (SAR) images~\cite{wang15,brunner10}, and Light Detection and Ranging (LiDAR) data~\cite{sampath2010segmentation,lidarradar18}. 
Such data, however, are expensive to obtain and hence the above approach is difficult to apply at a large scale, e.g., to all the buildings on earth. Moreover, such data is usually proprietary and not available to the public and research community. 
Recently, street scene images (or together with 2D maps) have been used for building height estimation~\cite{Yuan2016,diaz2016algorithm}, which can be easily obtained at large scale (e.g., via open source mapping applications such as Google Street View~\cite{Anguelov2010} and OpenStreetMap~\cite{Haklay2008}). 
Estimating building height via street scene images relies on accurate detection of building rooflines from the images, which then enables building height computation using camera projection. 
However, existing methods for roofline detection check the roofline segments only, which may be confused with overlapping buildings.
As shown in Fig.~\ref{fig:illustration}, the roofline of building B may be detected as the roofline of building A because the rooflines of different buildings may be in parallel with each other, and the buildings have similar colors and positions in the street scene images.

In this paper, we present a novel algorithm named Corner-based Height Estimation (\textbf{CBHE}) to estimate building height for \textit{complex street scene images} (with blocking problem from other buildings and trees). Our key idea to handle overlapping buildings is to detect not only the rooflines but also building corners. We obtain coordinates of building corners from building footprints in a 2D map (e.g., OpenStreetMap). We then map the corner coordinates into the street scene image to detect building corners in the image. Corners of different buildings do not share the same coordinates, and it is easier to associate them with different buildings, as shown in Fig.~\ref{fig:illustration}.

CBHE works as follows. It starts with an upper-bound estimation of the height of a building, which is computed as the maximum height that can be captured by the camera. It then tries to locate a line (i.e., a roofline candidate) at this height and repeats this process by iteratively reducing the height estimation. Following a similar procedure, CBHE also locates a set of building corner candidates. Next, CBHE filters the roofline candidates with the help of the corner candidates (i.e., a roofline candidate needs to be connected to a corner of the same building to be a true roofline). When the true roofline is identified, CBHE uses the pinhole camera model to compute the building height.

In the process above, when locating the roofline and corner candidates, we fetch two sets of image segments that may contain building rooflines or corners, respectively. To filter each set of objects and identify the true roofline and corner images, we propose a deep neural network model named \textbf{BuildingNet}. The key idea of BuildingNet is as follows. Building corner has a limited number of patterns (e.g., " \includegraphics[width=0.1in]{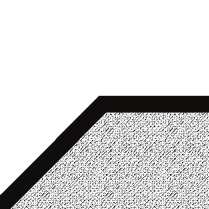}", "\includegraphics[width=0.1in]{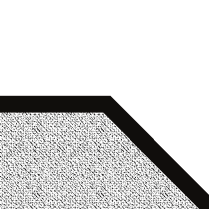}", "\includegraphics[width=0.05in]{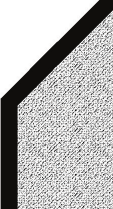}", "\includegraphics[width=0.05in]{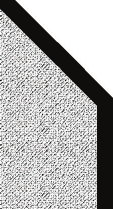}"), while non-corner images may have any pattern. The same applies to the building rooflines. Thus, we model building corner (roofline) identification as an \textit{open set} image classification problem, where each corner (roofline) pattern forms a class, while non-corner (non-roofline) images should be differentiated from them when building the classifier. To do so, BuildingNet learns embeddings of the input images, which minimize the intra-class distance and maximize the inter-class distance, and then differentiates different classes using a Support Vector Classification (SVC) model on the learned embeddings. When a new image comes, the trained SVC model will tell whether it falls into any corner (roofline) classes. If the image does not fall into any corner (roofline) classes, it is a non-corner image and can be safely discarded.

When estimating building height via the pinhole camera model, the result highly relies on the accuracy of the camera location due to GPS errors. Therefore, CBHE calibrates the camera location before roofline detection. To calibrate the camera location, CBHE detects candidate regions of all building corners in street scene images by matching buildings in street scene images with their footprints in 2D maps based on the imprecise camera location from GPS. Then, it uses BuildingNet to classify the corner candidates and remove those images classified as non-corner. From the remaining corner after the classification through BuildingNet, CBHE selects two corners with the highest score (detailed in Section~\ref{sec:camLoc_corDec}) to calibrate the camera location via the pinhole camera model. We summarize our contributions as follows:

\begin{itemize}[leftmargin=*]
	\item We model building corner and roofline detection as an open set classification problem and propose a novel deep neural network named BuildingNet to solve it. BuildingNet learns embeddings of the input images, which minimize the intra-class distance and maximize the inter-class distance. Experimental results show that BuildingNet achieves higher accuracy on building corner and roofline identification compared with the state-of-the-art open set classifiers.
	\item We propose a corner-based building height estimation algorithm named CBHE, which uses an entropy-based algorithm to select the roofline among all candidates from BuildingNet. The entropy-based algorithm considers both building corner and roofline features and yields higher robustness for the overlapping problem in complex street scene images. Experiments on real-world datasets show that CBHE outperforms the baseline method by over 10\% regarding the estimation error within two meters.
	\item We propose a camera location calibration method with an analytical solution when given the locations of two building corners in a 2D map, which means highly accurate result can be guaranteed with the valid building corners from BuildingNet.
\end{itemize}

We organize the rest of this paper as follows. We review related work in Section~\ref{sec:relatedWork} and give an overview of CBHE in Section~\ref{sec:preliminary}. The BuildingNet and entropy-based ranking algorithm are presented in Section~\ref{sec:camLoc}, and the building height estimation method is detailed in Section~\ref{sec:heightEst}. We report experimental results in Section~\ref{sec:experiment} and conclude the paper in Section~\ref{sec:conclusion}.

\section{Related Work}
\label{sec:relatedWork}
In this section, we review studies on camera location calibration and building heigh estimation. We also detail our baseline method~\cite{Yuan2016}.

\subsection{Camera Location calibration}
Camera location calibration aims to refine the camera location of the taken images, given rough camera position information from GPS devices or image localization~\cite{agarwal2015metric,liu2017efficient}.

Existing work uses 2.5D maps (2D maps with height information) to calibrate camera locations. Arth et al.~\cite{Arth2015} present a mobile device localization method that calibrates the camera location by matching building facades in street scene images with their footprints in 2.5D maps. Armagan et al.~\cite{Armagan2017} train a \emph{convolutional neural network} (CNN) to predict the camera location based on a semantic segmentation of the building facades in input images. Their method iteratively applies CNN to compute the camera's position and orientation until it converges to a location that yields the best match between building facades and 2.5D maps. Camera location calibration using 2.5D maps can produce good results. The hurdle is the requirement of building height information for generating 2.5D maps, which may not be available for every building.
Chu et al.~\cite{chu2014gps} extract the position features of \textit{building corner lines} (the vertical line of a corner) and then find the camera location and orientation by matching the extracted position features with building footprints in 2D maps. However, their method cannot handle buildings overlapping with each other or having non-uniform patterns on their facades.

\begin{figure*}[tp!]
	\centering
	\setlength{\abovecaptionskip}{1pt}   
	\setlength{\belowcaptionskip}{-2pt}   
	\includegraphics[width=0.9\linewidth]{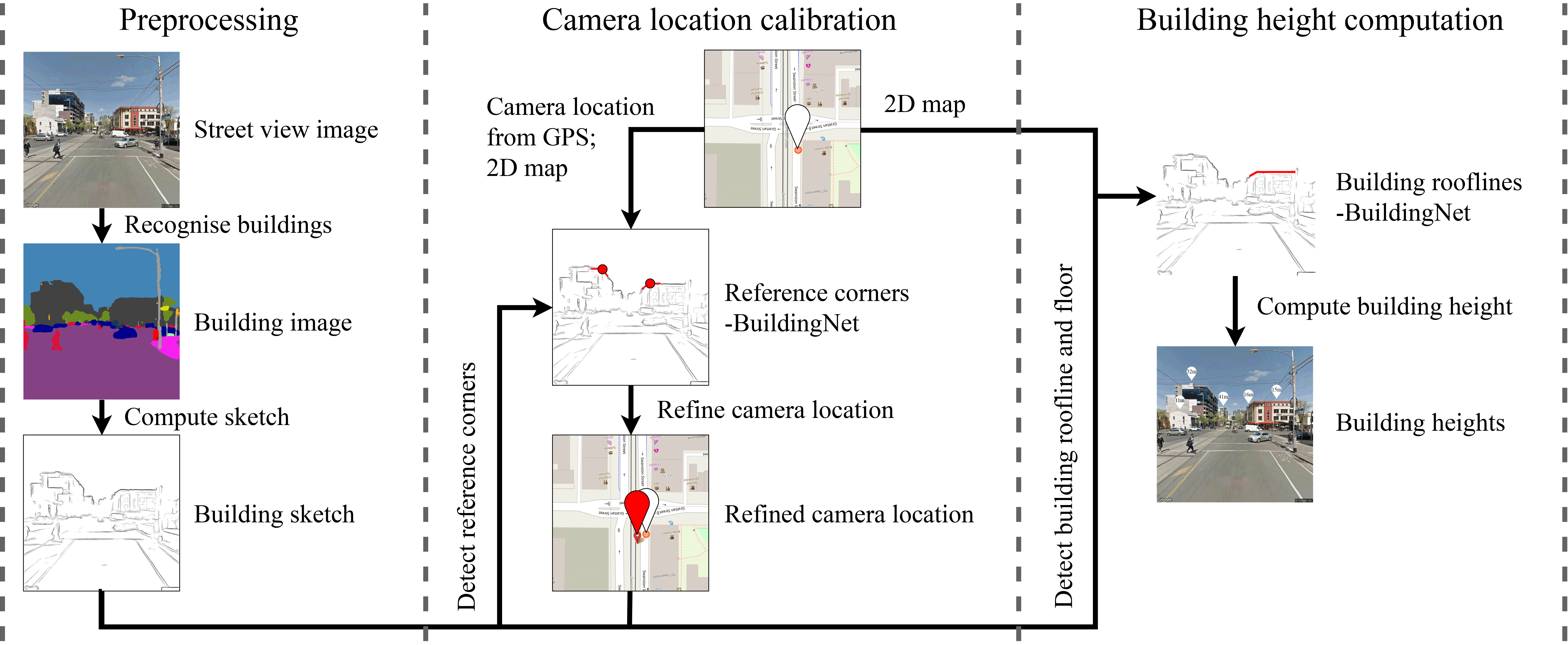}
	\caption{Solution overview}
	\label{fig:solution_framework}
\end{figure*}

\subsection{Building Height Estimation}
Building height estimation has been studied using geographical data such as high-resolution images, \textit{synthetic aperture radar} (SAR) images, and \textit{Light Detection and Ranging} (LiDAR) data. 

Studies~\cite{liasis16,izadi2012,zeng2014,tack12,qi2016building} based on high-resolution images (such as satellite or optical stereo images) estimate building height via methods such as elevation comparison and shadow detection, which may be impacted by lighting and weather condition when the images are taken. 
Similarly, methods based on height estimation is synthetic aperture radar (SAR) images~\cite{wang15,brunner10,sportouche2011extraction} are mainly based on shadow or layover analysis. Methods based on aerial images and aerial LiDAR data~\cite{sohn2008using,sampath2010segmentation} usually segment, cluster and then reconstruct building rooftop planar patches according to predefined geometric structures or shapes~\cite{zeng2014}. LiDAR data is expensive to analysis and has a limited operating altitude because the pulses are only effective between 500 and 2,000 meters~\cite{lidarradar18}. 
A common limitation shared by the methods above is that the data that they use are expensive to collect, which significantly constraints the scalability of these methods. 

\textbf{Method based on street scene image and 2D map.} Yuan and Cheriyadat propose a method for building height estimation uses street scene images facilitated by 2D maps~\cite{Yuan2016}. Street scene images are widely available from Google Street View~\cite{Anguelov2010}, Bing StreetSide~\cite{kopf2010street} and Baidu Map~\cite{baidu18}, 
which makes building height estimation based on such data easier to scale. Yuan and Cheriyadat's method has four main steps: (i) Match buildings in a street scene image with their footprints in a 2D map via camera projection based on the camera location that comes with the image. Here, the camera location may be imprecise due to GPS error \cite{zandbergen11,grammenos2018you}. (ii) Calibrate the camera location via camera projection with the extracted building corner lines from street scene images. (iii) Re-match buildings from a 2D map with those in the street scene image based on the calibrated camera location, and then detect building rooflines through edge detection methods. (iv) Compute building height via camera projection with camera parameters, calibrated camera location, the height of building rooflines in the street scene image, and the building footprint in the 2D map.

Our proposed CBHE differs from Yuan and Cheriyadat's method in the following two aspects: (A) In Step (ii) of their method, they calibrate camera location by matching building corner lines in the street scene image with building footprints in the 2D map.
Such a method cannot handle images in urban areas where the corner lines of different buildings are too close to be differentiated, or the buildings have non-uniform patterns/colors on their facades which makes corner lines difficult to recognize. CBHE uses building corners instead of corner lines, which puts more restriction on the references for camera location calibration, and thus yields more accurate results. (B) In Step (iv) of their method, they use a local spectral histogram representation~\cite{liu2002spectral} as the edge indicator to capture building rooflines, which can be ineffective when buildings overlap with each other. CBHE uses the proposed deep neural network named BuildingNet to learn a latent representation of building rooflines, which has been shown to be more accurate in the experiments. 

\section{Overview of CBHE}
\label{sec:preliminary}

We present the overall procedure of our proposed CBHE in this section. We also briefly present the process of camera projection, which forms the theoretical foundation of building height estimation using street scene images.

\subsection{Solution Overview}\label{sec:framework}
We assume a given street scene image of buildings that comes with geo-coordinates and angle of the camera by which the image is taken. Here, the geo-coordinates may be imprecise due to GPS errors. Google Street View images are examples of such images, and we aim to compute the height of each building in the image. As illustrated in Fig.~\ref{fig:solution_framework}, CBHE contains three stages:

\begin{itemize}[leftmargin=*]
	\item \textbf{Stage 1 -- Preprocessing:} In the first stage, we pre-process the input image by recognizing the buildings and computing their sketches. There are many methods for these purposes. We use two existing models RefineNet~\cite{Lin2017} and Structured Forest~\cite{dollar2013structured} to identify the buildings and compute their sketches, respectively. After this step, the input image will be converted into a grayscale image with each pixel valued from 0 to 255 that contains building sketches, which enables identifying rooflines and computing the height of the building via camera projection. 
	
	\item \textbf{Stage 2 -- Camera location calibration:} 
	Before computing building height by camera projection, in the second stage, we calibrate the camera location. This is necessary because a precise camera location is required in the camera projection, while the geo-coordinates that come with street scene images are imprecise due to GPS errors. To calibrate the camera location, we first detect building corner candidates in street scene images according to their footprints in 2D maps and their relative position to the camera. Then, by comparing the locations and the projected positions of building corners (two corners), we calibrate the camera location via camera projection. In this stage, we propose a deep neural network named BuildingNet to determine whether an image segment contains a valid building corner. The BuildingNet model and the process of selecting two building corners for the calibration are detailed in Section~\ref{sec:camLoc}.
	
	\item \textbf{Stage 3 -- Building height computation:} In this stage, we obtain the roofline candidates of each building via weighted Hough transform and filter out those invalid roofline candidates via BuildingNet. Then we rank all valid rooflines by an entropy-based ranking algorithm considering both corner and roofline features and select the best one for computing building height via camera projection. The detailed process is provided in Section~\ref{sec:heightEst}.
\end{itemize}

Since Stage 1 is relatively straightforward, we focus on Stages 2 and 3 in the following Sections~\ref{sec:camLoc} and~\ref{sec:heightEst}, respectively. Before diving into these two stages, we briefly discuss the idea of camera projection and present the frequently used symbols.

\subsection{Camera projection}
\begin{figure}[tp]
	\centering
	\setlength{\abovecaptionskip}{1pt}   
	\setlength{\belowcaptionskip}{-2pt}   
	\includegraphics[width=0.95\linewidth]{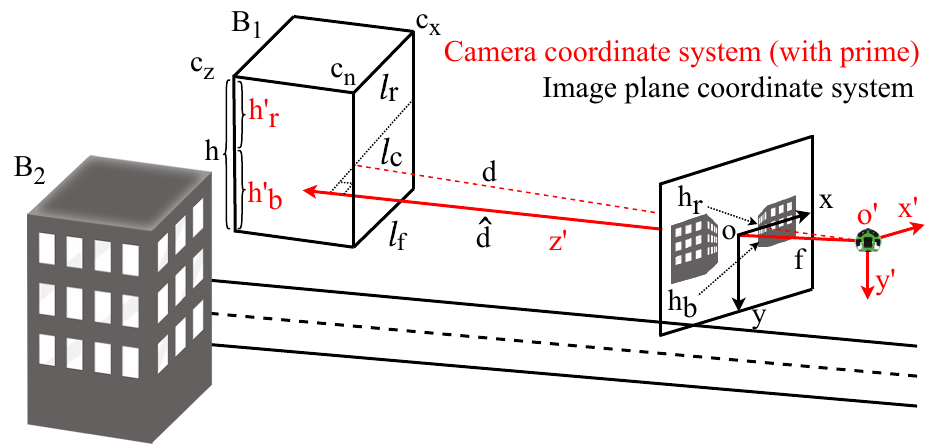}
	\caption{Geometric variables in the camera and the image coordinate systems (best view in color).}
	\label{fig:camera_projection}
\end{figure}

We use Fig.~\ref{fig:camera_projection} to illustrate the idea of camera projection and the corresponding symbols. 
In this figure, there are two coordinate systems, i.e., the camera coordinate system and the image plane coordinate system. Specifically, $\{o'$, $x'$, $y'$, $z'\}$ represent the camera coordinate system, where origin $o'$ represents the location of the camera. The camera is set horizontal to the sea level, which means that plane $x'z'$ is vertical to the building facades while the $y'$-axis is horizontal to the building facades. We use $\{o$, $x$, $y\}$ to represent the image plane coordinate system, where origin $o$ is the center of the image, and plane $xy$ is parallel to plane $x'y'$.

In Fig.~\ref{fig:camera_projection}, there are two buildings $B_1$ and $B_2$ that have been projected onto the image. For each building, we use $l_r$, $l_f$, and $l_c$ to represent the roofline, the floor, and the line on the building projected to the $x$-axis (center line) of the image plane $xy$. Corners $c_n$, $c_x$, and $c_z$ are the corner nearest to the camera, the corner farthest to the $y$-axis of the image plane when projected to the image plane (along the $x$-axis), and the corner closest to the $y$-axis of the image plane when projected to the image plane (along the $z$-axis), respectively. The height $h$ of the building is the sum of the distance between $l_r$ and $l_c$ and the distance between $l_c$ and $l_b$. These two distances are denoted as $h'_r$ and $h'_b$, and the projected length of $h'_r$ in the image plane $xy$ is denoted by $h_r$.
Since the camera is set horizontal to the sea level, the height of $h'_b$ is the same as the height of the car or human beings who captured the street scene image, which can be regarded as a constant. 

Let $d$ be the distance from the camera $o'$ to corner $c_n$, $\hat{d}$ be the projected length of $d$ onto the $z'$-axis, and $f$ be the focal length of the camera (i.e., the distance between the image center $o$ and the camera center $o'$). Based on the \emph{pinhole camera projection}, the height of a building can be computed as follows:

\begin{equation} 
h = h'_r + h'_b
= h_r \cdot \hat{d} / f + h'_b
\label{for:h}
\end{equation}

In this equation, the focal length $f$ comes with the input image as its metadata. The distance $\hat{d}$ is computed based on the geo-coordinates of the building and the camera, as well as the orientation of the camera. The geo-coordinates of the building are obtained from an open-sourced digital map, OpenStreetMap, while the geo-coordinates and orientation of the camera come with the input image from Google Street View. Due to GPS errors, we describe how to calibrate the location of the camera in Section~\ref{sec:camLoc}. The height $h_r$ is computed based on the position of the roofline $l_r$ which is discussed in Section~\ref{sec:heightEst}.
Table~\ref{tab:notation} summarizes the symbols that are frequently used in the rest of the discussion.

\begin{table}[tb]
	\setlength{\abovecaptionskip}{4pt}   
	\caption{Frequently used symbols}
	\label{tab:notation}
	\centering
	\begin{tabular}{cp{6.5cm}}
		\toprule
		\textbf{Notation} &\textbf{Description} \\
		\midrule
		$h_r$ & the height of a building above images' center line \\
		$h_b$ & the height of a building below images' center line \\
		$d$ & the distance from the camera to $c_n$ of a building\\
		$\hat{d}$ & the projected length of $d$ onto the $z$-axis \\
		$f$ & the focal length of the camera \\
		$l_r$ & a building roofline \\
		$c_n$ &    the corner nearest to the camera\\
		$c_x$ &     the corner farthest to $o$ in the image plane\\
		$c_z$ &    the corner closest to $o$ in the image plane\\
		\bottomrule
	\end{tabular}
\end{table}

\section{Camera Location Calibration}
\label{sec:camLoc}
When applying camera projection for building height estimation, we need the distance $\hat{d}$ between the building and the camera. Computing this distance is based on the locations of both the building and the camera. Due to the error of GPS, we calibrate the camera location in this section. 

\subsection{Key Idea}
We use two building corners in the street scene image with known real-world locations for camera location calibration. To illustrate the process, we project Fig.~\ref{fig:camera_projection} to a 2D plane, as shown in Fig.~\ref{fig:4}a, and assume that corner $c_n$ of building $B_1$ and building $B_2$ are two reference corners. 

We consider a coordinate system with corner $c_n$ of building $B_1$ as the origin, and the camera orientation as the $y$-axis. 
Let $\theta_1$ and $\theta_2$ be the angles of corner $c_n$ of $B_1$ and corner $c_n$ of $B_2$ from the orientation of the camera, respectively. Then the ratio of $d_2/d_3$ is determined by the position of these two reference corners in the image.
$\theta_3$ represents the angle between the line connecting corner $c_n$ of $B_1$ and corner $c_n$ of $B_2$ and $x$-axis, and it can be computed according to the camera's orientation and the relative locations of the two reference corners in 2D maps. Therefore, we can compute the coordinates $(x,y)$ of corner $c_n$ of building $B_2$ in the coordinate system. With $\theta_1$, $\theta_2$, and the coordinate $(x,y)$, we compute the $y$ coordinate of the camera as follows:

\begin{equation} 
\begin{aligned}
y' = &\dfrac{x-y\cdot tan\theta_1}{tan\theta_1+tan\theta_2}
\label{for:cl}
\end{aligned}
\end{equation}

Since the $x$ coordinate of the camera is equals to $y'\cdot tan\theta_2$, we obtain the relative position of the camera to the corner $c_n$ of building $B_1$.
Thus, camera location calibration becomes the problem of matching two building corners with their positions in the image.

The real-world location of the building corners can be obtained from 2D maps, and we need to locate their corresponding positions in the street scene image based on the (inaccurate) geo-coordinates of the camera. For a pinhole camera, matching a 3D point in the real world to a 2D point in the image is determined by a $3{\times}4$ camera projection matrix as follows:

\begin{figure}[tp]
	\centering
	\setlength{\abovecaptionskip}{1pt}   
	\setlength{\belowcaptionskip}{-2pt}   
	\begin{subfigure}[a]{0.42\linewidth}
		\includegraphics[width=\linewidth]{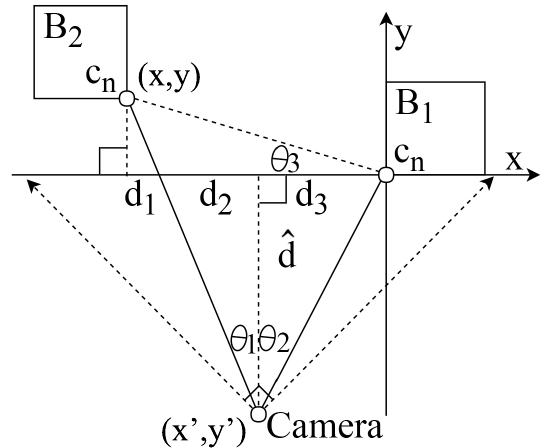}
		\caption{}
	\end{subfigure}\;\;\;\;\;
	\begin{subfigure}[a]{0.333\linewidth}
		\includegraphics[width=\linewidth]{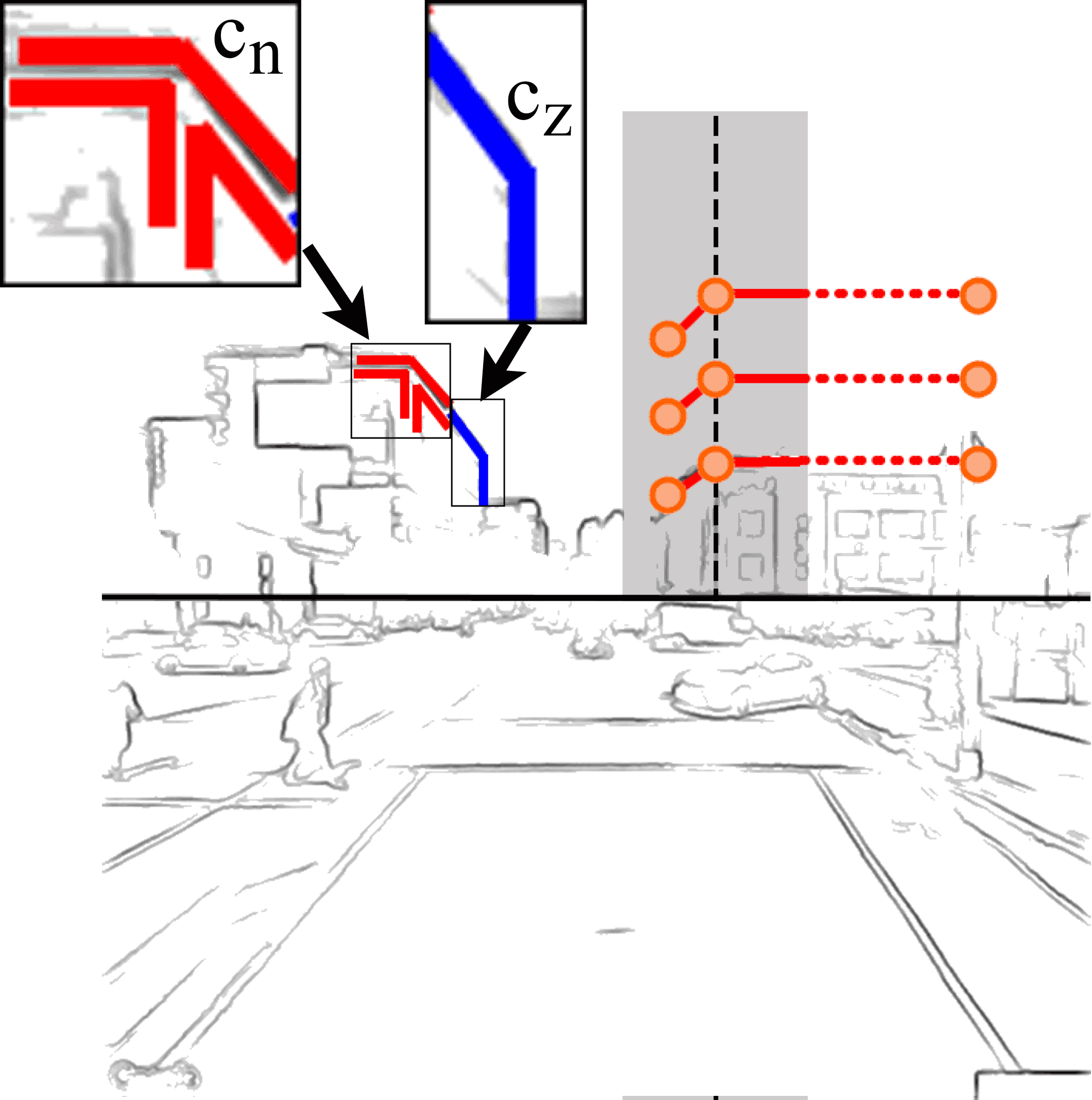}
		\caption{}
	\end{subfigure}
	\caption{(a) geometric variables of Fig.~\ref{fig:camera_projection} in plan view. (b) the left building shows the formation of $c_n$ and $c_z$, while the right building illustrates how to find corner candidates.}
	\label{fig:4}
\end{figure}

\begin{equation} 
{\alpha\cdot p} = [I|0_3] \begin{bmatrix}R & t \\ 0_3^T & 1\end{bmatrix}\left[\begin{array}{c} p' \\ 1 \end{array} \right] , \; I = \begin{bmatrix} f&0&0 \\ 0&f&0 \\ 0&0&1\end{bmatrix}
\label{for:p}
\end{equation}
where a real-world point $p'$ = $(x',y',z')^T$ can be projected to its position $p$ = $(x,y,1)^T$ in the image plane; $\alpha$ is the parameter that transfers pixel scale to millimeter scale~\cite{meyer10}; $[I|0_3]$ is the camera matrix determined by focal length $f$; $R$ is the camera rotation matrix, while $t$ is a 3-dimensional translation vector that describes the transformation from the real-world coordinates to the camera coordinates. 

Since the image geo-coordinates may be inaccurate, we can only compute rough locations of the building corners. Based on the rough position of each corner, we then iteratively assume a height $h_r$ for each building to obtain the gradient of its rooflines, as shown in Fig.~\ref{fig:4}b. We use $120\times120$ sub-images with the horizontal position and the assumed height of the corner as their center for building corner detection. A building corner consists of two rooflines or a roofline with a building corner line, as shown in Fig.~\ref{fig:4}b. For each building, we only consider their corners $c_n$ and $c_z$. There are three types of formation for corner $c_n$ as illustrated by the red lines on the left-hand side building in Fig.~\ref{fig:4}b, and there is one type of formation for corner $c_z$ as illustrated by the blue lines. Based on the detected building corner candidates, we use BuildingNet described Section~\ref{sec:camLoc_corDec} to filter out non-corner image segments, and then select the two reference corners which is discussed in Section~\ref{sec:camLoc_corDec}.

We assume the camera location error from Google Street View API to be less than three meters due to its camera location optimization~\cite{klingner2013street}. If the camera location we compute is more than three meters away from the one provided by Google Street View API, we use the camera location from Google Street View API directly. We further improve the estimation accuracy by a multi-sampling strategy, which uses the median height among results from different images of the same building taken at different distances.

\subsection{BuildingNet}
\label{sec:camLoc_corDec}

We formulate building corner detection as an object classification problem, which first detects candidate corner regions for a specific building by a heuristic method, and then classifies them into different types of corners or non-corners. 

We classify images that may contain building corners into five classes. The first four classes correspond to images containing one of the four types of building corners, i.e., corner $c_n$, $c_z$ of the left-hand side buildings, and corner $c_n$, $c_z$ of the right-hand side buildings (" \includegraphics[width=0.1in]{./figure/t1.eps}", "\includegraphics[width=0.1in]{./figure/t2.eps}", "\includegraphics[width=0.05in]{./figure/t4.eps}", "\includegraphics[width=0.05in]{./figure/t3.eps}"). 
The last class corresponds to non-corner images which may contain any pattern except the above four types of corners (e.g., they could contain trees, lamps or power lines), and should be filtered out. Such a classification problem is an \textit{open set} problem in the sense that the non-corner images do not have a unified pattern and will encounter unseen patterns.
To solve this classification problem, we build a classifier that only requires samples of the first four classes in the training stage (can also take advantage of non-corner images), while can handle all five classes in the testing stage.
To enable such a classifier, we first propose the \emph{BuildingNet} model based on LeNet 5~\cite{lecun1998gradient} and triplet loss functions, which learns embeddings that map potential corner region image segments to a Euclidean space where the embeddings have small intra-class distances and large inter-class distances.


\subsubsection{Triplet Relative Loss Function}
\label{sec:loss}
\begin{figure}[tp]
	\centering
	\setlength{\abovecaptionskip}{1pt}   
	\setlength{\belowcaptionskip}{-2pt}   
	\includegraphics[width=0.95\linewidth]{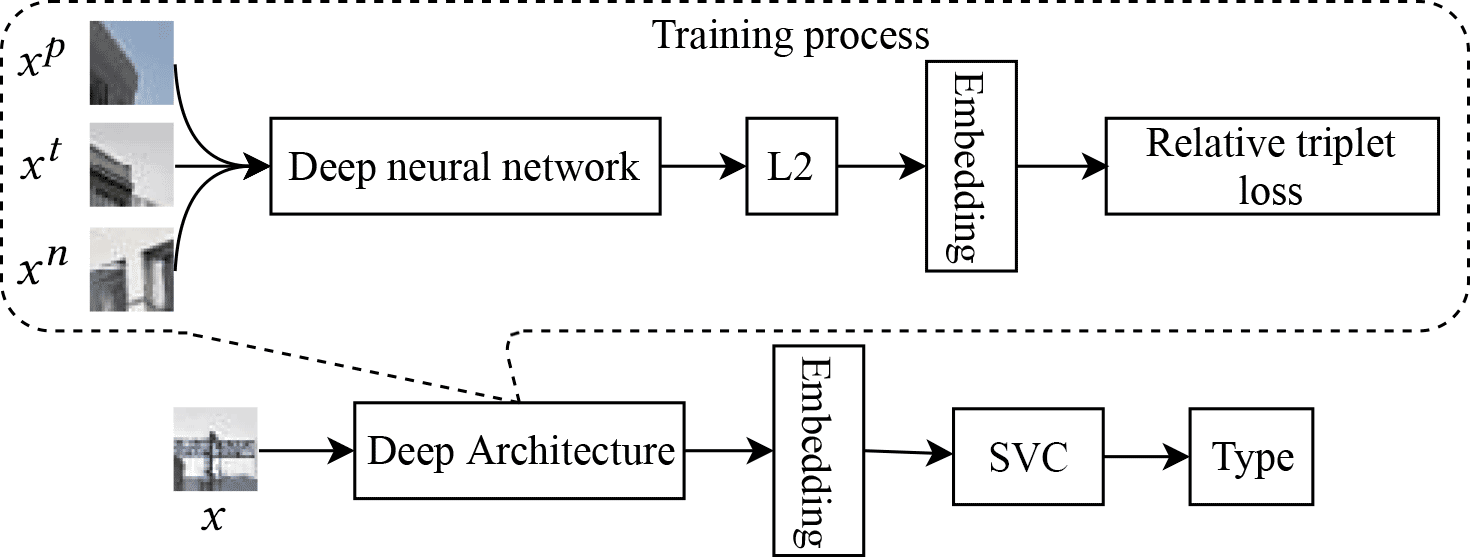}
	\caption{BuildingNet structure. $x^t$, $x^p$ are images containing the same corner type. $x^n$ is an image containing another corner type or non-corner, $x$ is a testing image.}
	\label{fig:4.1t}
\end{figure}

As shown in Fig~\ref{fig:4.1t}, an input of BuildingNet contains three images. Two of them ($x^p$ and $x^t$) contain the same type of corner, and we name them the target ($x^t$) and positive ($x^p$), respectively. The other image $x^n$ contains another type of corners (or a non-corner image if available), and we name it negative. 
BuildingNet trains its inputs to $d$-dimensional embeddings based on a \textit{triplet relative loss} function inspired by Triplet-Center Loss and FaceNet~\cite{schroff2015facenet,wen2016discriminative,he2018triplet,wang2018kdgan}, which minimizes the distances within the same type of corners, and maximizes the distances between different types of corners as follows:

\begin{eqnarray} 
l = \!\sum_{i=1}^{N} \alpha \cdot || f(x_i^t)-f(x_i^p)||_2^2 + (1-\alpha) \cdot \dfrac{||f(x_i^t)-f(x_i^p)||_2^2}{||f(x_i^t)-f(x_i^n)||_2^2} \!
\label{for:tl}
\end{eqnarray}
where $\alpha\in[0,1]$ is the weight of intra-class distance in the $d$-dimensional Euclidean space;
$(1-\alpha)$ is the weight of the ratio between intra-classes distance and inter-class distance, which aims to separate different classes in the $d$-dimensional Euclidean space; $N$ is the cardinality of all input triplets. Function $f$ computes the $d$-dimensional embedding of an input image, and we normalize it to $||f(x)||_2^2 = 1$. Different from existing loss function based on triplet selection~\cite{weinberger2006distance,schroff2015facenet}, \textit{triplet relative loss} function can minimise the intra-class distance and maximize the inter-class distance by means of their relative distance.

\subsubsection{Hard Triplet Selection}
\label{sec:hard_triplet}

Generating all possible image triplets for each batch during the training process will result in a large amount of unnecessary training data (e.g., $x^t$ and $x^p$ are too similar, while $x^n$ is way different). It is crucial to select triplets that contribute more to the training phase. In BuildingNet, we accelerate convergence by assigning a higher selection probability to triplets that may contribute more to the training process. The probability of selecting a negative image $x_i^n$ to form a training triplet is:

\begin{eqnarray} 
\begin{aligned}
p(x_i^n)&=\!\dfrac{e^{m-||f(x_i^n)-f(x^t)||_2^2}}{\sum_{i=1}^{k}e^{m-||f(x_i^n)-f(x^t)||_2^2}},\; i = [1, k]\!\\
m&=\!min(||f(x_i^n)-f(x^t)||_2^2-||f(x^t)-f(x^p)||_2^2),\; i = [1, k]\!
\label{for:lw}
\end{aligned}
\end{eqnarray}

Here, $k$ is the total number of negative images in a batch. After randomly choosing $x^t$ and $x^p$ for a triplet, we compute the Euclidean distance between $x^t$ and $x^p$, as well as the distances between $x^t$ and the $k$ negative images $x^n$ in the batch. Let $m$ be the minimum Euclidean distance between $x^t$ and any $x^n$, which can be positive or negative. Then, the negative image $x_i^n$ similar to $x^t$ will have a higher probability to be selected.

After the training process, we obtain a $d$-dimensional embedding for each input image. We then learn a support vector classifier~\cite{chang2011libsvm} based on these embeddings for corner region image classification. 



\subsection{Entropy-based Ranking}
\label{sec:camLoc_roofWei}

BuildingNet can filter out non-corner images. Among the remaining corner candidates, we select the two corners with the highest score as the reference corners. Reference corner selection relies on multiple factors: the length and edgeness (detailed in Section~\ref{sec:heightEst}) of the lines forming the corner, the number of other corner candidates ($c_n$, $c_x$, and $c_z$) of the same building with the same assumed height, and the position of the corner in the image. We take the position of the corner into consideration because, empirically, corners close to one quarter or three-quarters (horizontally) of the image yield more accurate matching between their positions in the image and their footprints in 2D maps. We also consider their real-world locations because a corner close to the camera will be clearer and has higher accuracy when matching them to their footprints in 2D maps. Therefore, we define the score of each corner candidate as:

\begin{equation} 
[s_{c_1},...,s_{c_k}]' = \begin{bmatrix} \lambda,\omega,\tau,\rho,d | c_1 \\ \vdots\\ \lambda,\omega,\tau,\rho,d | c_k \end{bmatrix}\cdot\begin{bmatrix} w_\lambda \\ \vdots \\ w_d\end{bmatrix}
\label{for:wc}
\end{equation}
where $k$ is the number of corner candidates from all buildings; $c_i$ is the $i$th corner candidate; $s_{c_i}$ is the score of the $i$th corner candidate; $\lambda$ is the detected length of the two lines that form a corner, while $\omega$ is the sum of the edgeness of the two lines; $\tau$ is the number of other corner candidates of the same building with the same assumed height; $\rho$ is the minimum distance from the corner to a quarter or three-quarters of the image, and $d$ is the distance from the corner to the camera; $w_\lambda$, $w_\omega$, $w_\tau$, $w_\rho$, $w_d$ are the weight of these parameters. Parameters $\lambda, \omega, \tau$ and $\rho$ correlate with the score positively, while parameter $d$ correlates with the score negatively. 

We use an entropy-based ranking method to compute the weights of parameters $(w_\lambda, ..., w_d)^T$.
Shannon entropy is a commonly used measurement of uncertainty in information theory~\cite{sun2017ecological}. The main idea of the entropy-based ranking algorithm is to compute the objective weights of different parameters according to their data distribution. If the samples of a parameter vary greatly, the parameter should be considered as a more important feature. 

For building corner classification, there are $n=5$ parameters and $m=k$ samples. We denote the decision matrix as $r$, where $r_{ij}$ is the value of the $i$th sample under the $j$th parameter. Before applying the entropy-based ranking algorithm, we pre-process these parameters by Min-max scaling as follows:

\begin{equation} 
r_{ij} = \left\{
\begin{aligned}
&\!(r_{ij} - \min\limits_{j}(r_{ij})) / (\max\limits_{j}(r_{ij}) - \min\limits_{j}(r_{ij})), \; \text{iff positive} \!\\
&\!(r_{ij} - \min\limits_{j}(r_{ij})) / (\max\limits_{j}(r_{ij}) - \min\limits_{j}(r_{ij})) + 1, \; \text{iff negative} \!
\label{for:scal}
\end{aligned}
\right.
\end{equation}
where positive and negative mean that the $j$th parameter is positively/negatively correlated with the value of $r$. After Min-max scaling, the entropy of each parameter based on the normalized decision matrix $r'$ is defined as:

\begin{equation} 
e_j = -ln(m)^{-1}\cdot \sum_{j=1}^m r'_{ij}\cdot ln (r'_{ij}), \; r'_{ij} = r_{ij}/\sum_{j=1}^m r_{ij}
\label{for:ent}
\end{equation}
where $r'$ is the standardized $r$. Based on the entropy of each parameter, the weight of each parameter is computed by:

\begin{equation} 
w_j = (1-e_j)/(n-\sum_{j=1}^m e_j), \; j = [1, n]
\label{for:wj}
\end{equation}

After computing the weight of each parameter, we apply them to all corner candidates and rank all the candidates by their scores to obtain the best two as the reference corners.

\section{Roofline Detection}
\label{sec:heightEst}
Building height estimation requires detecting the roofline of each building. In this section, we present our method for roofline candidate detection in Section~\ref{sec:rcd}, and our method for the true roofline selection in Section~\ref{sec:rcr}. We further present a strategy for handling tall building (over 100 meters) in Section~\ref{sec:tbp}.

\subsection{Roofline Candidate Detection}
\label{sec:rcd}
We consider the rooflines from corner $c_n$ to the corner next to $c_n$, along the positive direction of the $x'$-axis in the camera coordinate system, and the one from corner $c_z$ to the corner next to corner $c_z$ along the negative direction of the $z'$-axis in the camera coordinate system.
The corner between corner $c_z$ and corner $c_x$ is corner $c_n$ if they are adjacent to each other, as shown in Fig.~\ref{fig:camera_projection}, and we take this situation to simplify the explanation. 

Similar to corner candidate detection, as shown in Fig.~\ref{fig:511}a, we find all roofline candidates of each building by a heuristic method, which projects the rooflines of each building according to its relative location to the camera in the real world, together with the camera's parameters. To do so,
we first assume $h_r$ of a building to be the maximum height that can be captured, which means that at least a roof corner ($c_n$, $c_x$, and $c_z$) is visible in the image. If corner $c_n$ is visible, the maximum height computed via camera projection is: 

\begin{figure}[tp]
	\centering
	\setlength{\abovecaptionskip}{1pt}   
	\setlength{\belowcaptionskip}{-2pt}   
	\begin{subfigure}[a]{0.333\linewidth}
		\includegraphics[width=\linewidth]{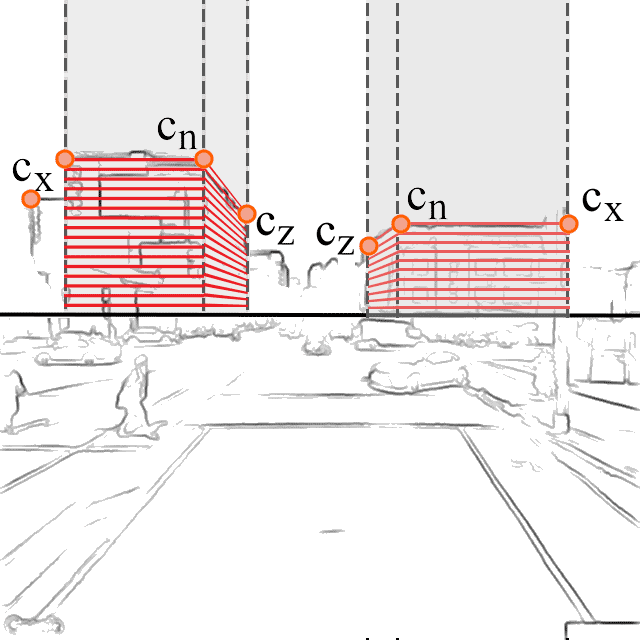}
		\caption{}
	\end{subfigure}\;\;\;\;\;\;
	\begin{subfigure}[a]{0.333\linewidth}
		\includegraphics[width=\linewidth]{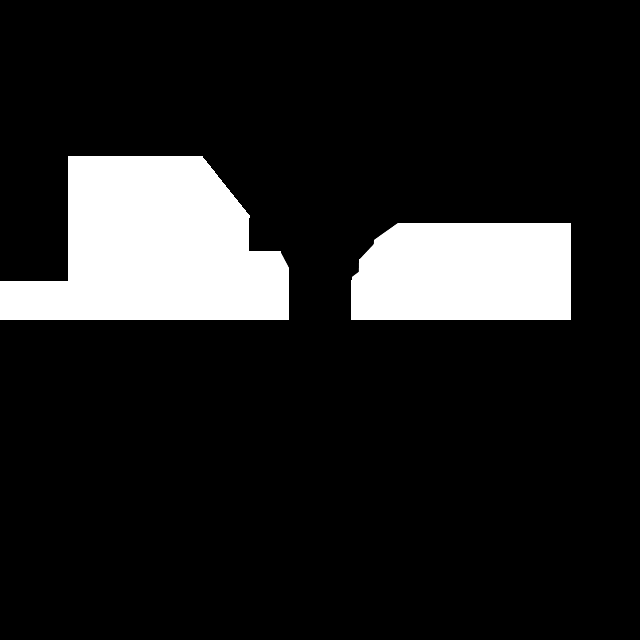}
		\caption{}
	\end{subfigure}
	\caption{(a) the heuristic method for roofline candidate detection. (b) the mask of detected buildings.}
	\label{fig:511}
\end{figure}

\begin{equation} 
h_r = \hat{d}\cdot(h_{I}/2f) 
\label{for:dd}
\end{equation}
where $h_{I}$ is the height of the street scene image; $\hat{d}$ is the distance from corner $c_n$ to the camera projected to the $z'$-axis of the camera coordinate system. If corner $c_n$ is invisible, we use $c_z$ as the reference corner when computing the maximum height of a building in the same way. With the maximum height of the building, we compute the position of corner $c_n$, $c_x$, and $c_z$ in the image. We then apply Hough transform to the input edge map in Fig.~\ref{fig:solution_framework} to detect roofline candidates, and the roofline candidates from $c_n$ to $c_x$ need to match the computed position of $c_n$ and $c_x$. Similarly, the roofline candidates from $c_n$ to $c_z$ need to match the computed position of $c_n$ and $c_z$. Instead of using the typical Hough transform for line detection, which takes binarized images as the input, we sum the value of all pixels valued from 0 to 255 within a line as its weight, and name the summed value as the \textbf{edgeness} of a roofline candidate, which reflects the visibility of a line in the edge map. 

We iteratively reduce the assumed height with a step length of 0.5 meters until $h_r = 0$ and collect all candidate rooflines. Similar to reference corner detection, we formulate the true building roofline detection as an open set classification and ranking problem.

\subsection{Roofline Classification and Ranking}
\label{sec:rcr}
There are three types of rooflines: (i) Roofline from $c_n$ to $c_x$; (ii) Roofline from $c_n$ to $c_z$ of the left hand side buildings; (iii) Roofline from $c_n$ to $c_z$ of the right hand side buildings, as shown in Fig.~\ref{fig:4}c. We use BuildingNet to filter these candidates and find the true roofline, which is similar to the corner candidate validation process in Section~\ref{sec:camLoc_corDec}.
Based on the valid roofline candidates from BuildingNet, we weight each roofline candidate $l_{r}$ by its detected length $\lambda$, edgeness $\omega$, and the number of corners $\tau$ with the same assumed height of the same building. We rank all roofline candidates via the entropy-based ranking algorithm in Section~\ref{sec:camLoc_roofWei}, as follows: 

\begin{algorithm}[tp]
	\setlength{\abovecaptionskip}{1pt}   
	\setlength{\belowcaptionskip}{-2pt}   
	\caption{Roofline pre-processing}
	\label{alg:pre-process}
	Inputs: buildings $B$, tree area $T$, edge map $E$\;
	Output: updated buildings $B$\;
	M = null\;
	\ForAll{building $b$ in $B$}{ 
		// buildings are ordered by whether they have a detectable corner, and then their distance to the camera\;
		\ForAll{roofline $l_r$ in $b.L_r$}{ 
			// traverse all roofline candidates of building $b$\;
			\ForAll{pixel $p \in l_r$}{ 
				\If{ $M(p)$}{
					$l_r$.delete(p)\;
				}
			}
			$l_{r(ini)} = l_r$\;
			\ForAll{pixel $p$ in $l_r$ and $p \notin l_r$ and $!M(p)$}{ 
				\If{$connected(p,l_r)$ and $T(p)$}{
					$l_r$.add(p)\;
				}
			}
			// update the length and edgeness of $l_r$\;
			$\lambda_{l_r} = len(l_r)$\;
			$\omega_{l_r} = E(l_{r(ini)}) * len(l_r)/len(l_{r(ini)})$\; 
		}
	}
\end{algorithm}

\begin{equation} 
[s_{l_{r_1}},...,s_{l_{r_k}}]' = \begin{bmatrix} \lambda,\omega,\tau | l_{r_1} \\ \vdots\\  \lambda,\omega,\tau | l_{r_k} \end{bmatrix}\cdot\begin{bmatrix} w_\lambda | l_r \\ \vdots \\ w_\tau | l_r \end{bmatrix}
\label{for:wl}
\end{equation}
where $k$ is the number of roofline candidates for a specific roofline; $l_{r_i}$ is the $i$th roofline candidate; $s_{l_{r_1}}$ is the score of the $i$th roofline candidate. $w_\lambda$, $w_\omega$ and $w_\tau$ are the weight of these parameters based on all candidates of a specific roofline of a building, and all three parameters are positively correlated with the score $s$. 
The value of $\tau$ depends on the number of corners ($c_n$, $c_x$, and $c_z$) with the same assumed height as the roofline candidate, and its value is \{0, 1, 2, 3\}. We discussed how to detect references corners in Section~\ref{sec:camLoc_corDec}, and the difference in detecting the corners of a specific building is that we do not consider $\rho$ and $d$ in Equation~\ref{for:wc} and all corner candidates here are those of a specific building corner. 

Different from building corners, which can only be visible or invisible, rooflines can also be partially blocked by other objects (trees in particular). Therefore, before we apply the ranking algorithm, we pre-process the length $\lambda$ and edgeness $\omega$ which are affected by the blocking via Algorithm~\ref{alg:pre-process} as follows:

When estimating building height, we first separate buildings into two classes: (i) with at least one valid corner; (ii) without any valid corner.  Then, we process buildings in class (i) according to their distance to the camera. After all the buildings in class (i) have been processed, we process the buildings in class (ii) according to their distance to the camera. After we obtain the height of a building, we mark the scope of the building in the street scene image, as shown in Fig.~\ref{fig:511}b (i.e., height has been obtained).

After detecting the roofline candidates of a building, we refine the $\lambda_{l_r}$ of each roofline candidate using the following equation:

\begin{equation} 
\lambda_{l_r} = \lambda_{l_r (ini)} - \sum_{p \in l_r} M(p) + \sum_{p\;in\;l_r} T(p)
\label{for:lr}
\end{equation}
where $\lambda_{l_r (ini)}$ is the detected length of a roofline candidate $l_r$.  $M(p)$ checks whether a pixel $p$ within a roofline belongs to a building's scope in the street scene image that has been processed and closer to the camera, or within another building's roofline that has been processed but farther to the camera. We remove pixel $p$ from a roofline if $M(p)$ is true. $T(p)$ checks whether a pixel $p$, which is in the extended line of $l_r$ but within the projected scope of the roofline, has been blocked by trees. If there do exist these pixels and they connect to the detected roofline segment, we add them to the roofline. Accordingly, we update the edgeness of a roofline as:

\begin{equation} 
\omega_{l_r} = (1+ \lambda_{l_r (ini)}/\lambda_{l_r})\cdot\sum_{p \in l_r} E(p)\cdot (1 - M(p))  
\label{for:er}
\end{equation}
where $E$ is the input edge map of the original image, $\lambda_{l_r (ini)}$ and $\lambda_{l_r}$ are the initial and prolonged length of the roofline, respectively.

\subsection{Tall Building Preprocessing}
\label{sec:tbp}
Height estimation for tall buildings (over 100 meters) requires the camera to be placed far away from the buildings with an upward-looking view to capture the building roof. For images with an upward-looking view, all building corner lines will become slanted. Typically, we take the upward-looking view as 25 degrees as an example to show the strategy that we use for handling tall buildings.

\begin{figure}[tp!]
	\centering
	\setlength{\abovecaptionskip}{1pt}   
	\setlength{\belowcaptionskip}{-2pt}   
	\begin{subfigure}[a]{0.333\linewidth}
		\includegraphics[width=\linewidth]{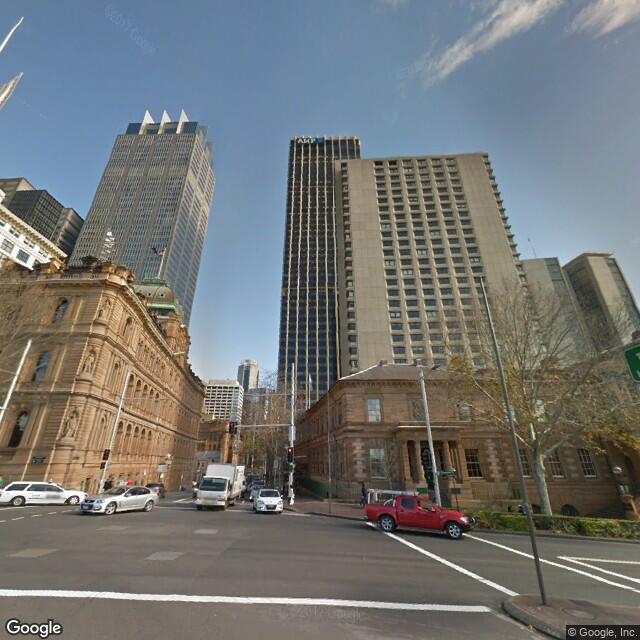}
		\caption{}
	\end{subfigure}\;\;\;\;\;\;
	\begin{subfigure}[a]{0.333\linewidth}
		\includegraphics[width=\linewidth]{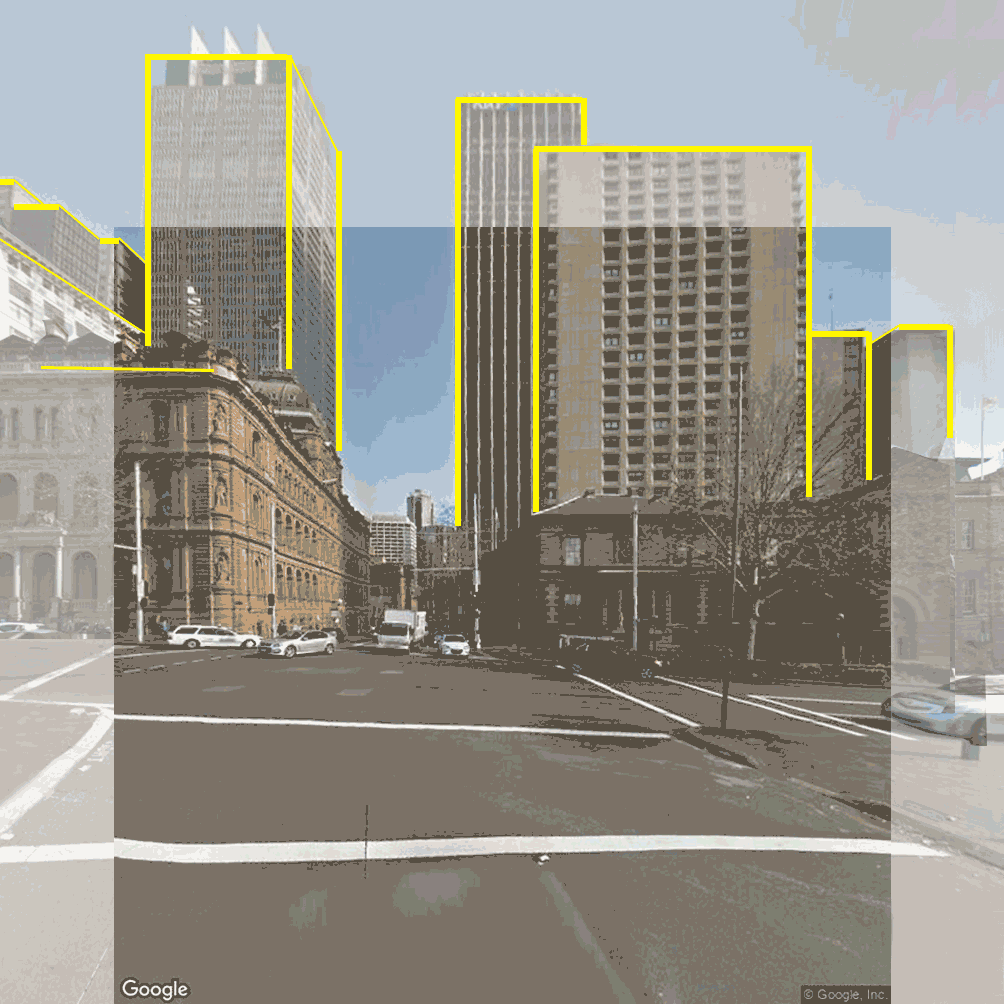}
		\caption{}
	\end{subfigure}
	\caption{(a) an image with an upward-looking view. (b) the corresponding image with a horizontal view.}
	\label{fig:53}
\end{figure}
We first compute a plane-to-plane homography~\cite{zhang2000flexible}, which maps an image with an upward-looking view to the corresponding image with a horizontal view. Here, we use the homogeneous estimation method~\cite{criminisi97}, which solves a $3{\times}3$ homogeneous matrix $h$ that matches a point in an upward-looking image (Fig.~\ref{fig:53}a) to a horizontal-view image (Fig.~\ref{fig:53}b) using the Equation~\ref{for:A}:

\begin{equation} 
A\cdot h =  \begin{bmatrix}\begin{smallmatrix}
x_1 & y_1 & 1 & 0 & 0 & 0 & -x_1X_1 & -y_1Y_1 & -X1 \\
0 & 0 & 0 & x_1 & y_1 & 1 & -x_1X_1 & -y_1Y_1 & -X1 \\
\vdots & \vdots & \vdots & \vdots & \vdots & \vdots & \vdots & \vdots & \vdots \\
x_n & y_n & 1 & 0 & 0 & 0 & -x_nX_n & -y_nY_n & -Xn \\
0 & 0 & 0 & x_n & y_n & 1 & -x_nX_n & -y_nY_n & -Xn \\
\end{smallmatrix}\end{bmatrix}\cdot h = 0
\label{for:A}
\end{equation}
where the homogeneous matrix $h (|h| = 1)$ is represented in the vector form as $h = (h_{11}, h_{12}, h_{13}, h_{21}, h_{22}, h_{23}, h_{31}, h_{32}, h_{33})^T$; $n$ is the number of point pairs, which should be no less than four to validate the homogeneous equation; ($X_i, Y_i$) represents a point in the upward-looking image and ($x_i, y_i$) represents the corresponding point in the resultant image with a horizontal view. Vector $h$ minimizes the algebraic residuals, and $A \cdot h$ is a standard result of linear algebra. Subject to $h = 1$, the least eigenvalue of $A\cdot A^T$ is given by the eigenvector, and this eigenvector can be obtained from the singular value decomposition (SVD) of $A$. 


\section{Experiments}
\label{sec:experiment}
In this section, we first evaluate our proposed BuildingNet model for building corner and roofline classification and then evaluate our proposed CBHE algorithm for building height estimation. 

\subsection{Datasets}
\label{sec:data}
In our experiments, we obtain building footprints (geo-coordinates) from OpenStreetMap and building images from Google Street View, respectively. For the experiments on building height estimation, we use two datasets: 

(i) \textbf{City Blocks}, which contains 128 buildings in San Francisco.
We collect all Google Street View images (640$\times$640 pixels) with camera orientation along the street.
We set the view of the camera is 90 degrees, and the focal length can be derived via the camera parameters provided by Google. We do not need to consider the camera rotation matrix $R$ and the translation vector $t$ due to the image preprocessing of Google Street View. We obtain the building height ground truth from high-resolution aerial maps (e.g., NearMap~\cite{nearmap18}).

(ii) \textbf{Tall Buildings}, which contains 37 buildings taller than 100 meters in San Francisco, Melbourne, and Sydney collected by us via Google Street View API. We set the camera with an upward-looking view (25 degrees) to capture their rooflines.
The building hight ground truth comes from Wikipedia pages of these buildings or derived from NearMap~\cite{nearmap18}.

\begin{figure}[tp]
	\centering
	\setlength{\abovecaptionskip}{1pt}   
	\setlength{\belowcaptionskip}{-2pt}   
	\includegraphics[width=\linewidth]{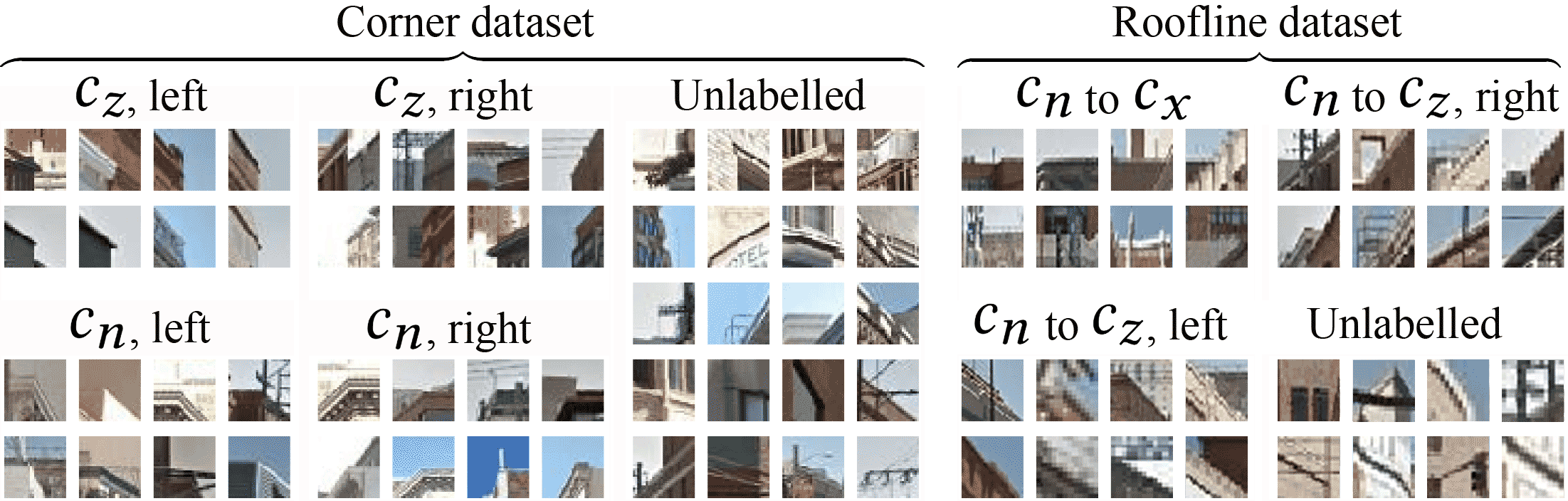}
	\caption{Examples of four types of corners, three types of rooflines, and the corresponding unlabelled images.}
	\label{fig:6.1}
\end{figure}

For building corner classification, we crop images from City Blocks dataset. We generate the corner dataset semi-automatically, where we crop $28\times28$ pixels image segments from street scene images, and then manually label whether an image segment contains a building corner (and the type of corner). The training dataset that we collected contains 10,400 images, including 1,300 images of each type of building corner (i.e., a total of 5,200 building corner images) and 5,200 non-corner images. The testing dataset contains 1,280 images, including 160 images for each type of building corners and 640 non-corner images. The training data and testing data come from different city blocks.

Following a similar approach, we collect a roofline dataset. For each roofline candidate, we extend the upper and lower 10 pixels of the roofline to obtain a $21\times W$ image segments, where $W$ is the length of the roofline, and we further resize (rotate if the roofline is not a horizontal line) the image to $28\times28$ to generate same-size inputs for BuildingNet. The training dataset includes 7,800 images, including 1,300 images for each type of rooflines (i.e., a total of 3,900 building roofline images) and 3,900 non-roofline images. The testing dataset contains 960 images, including 160 images for each kind of roofline and 480 non-roofline images.

\subsection{Effectiveness of BuildingNet}
\label{sec:effectivenessbuildingnet}
Building corner and roofline classification is an open set classification problem where the invalid corner or roofline candidates do not have consistent features. To test the effectiveness of BuildingNet, we use two open set classifiers as the baselines: \textit{SROSR}~\cite{zhang2017sparse} and \textit{OpenMax}~\cite{bendale2016towards}. SROSR uses the reconstruction errors for classification. It simplifies the open set classification problem into testing and analyzing a set of hypothesis based on the matched and no-matched error distributions. OpenMax handles the open set classification problem by estimating the probability of whether an input comes from unknown classes based on the last fully connected layer of a neural network. Further, we use two loss functions based on triplet selection to illustrate the effectiveness of our proposed triplet relative loss function. The loss function in \textit{FaceNet}~\cite{schroff2015facenet} makes the intra-class distance smaller than inter-class distance by adding a margin, and the one in \textit{MSML}~\cite{xiao2017margin} optimizes the triplet selection process towards selecting hard triplets in each in training.

\begin{figure}[tp]
	\centering
	\setlength{\abovecaptionskip}{1pt}   
	\setlength{\belowcaptionskip}{-2pt}   
	\begin{subfigure}[b]{0.49\linewidth}
		\includegraphics[width=\linewidth]{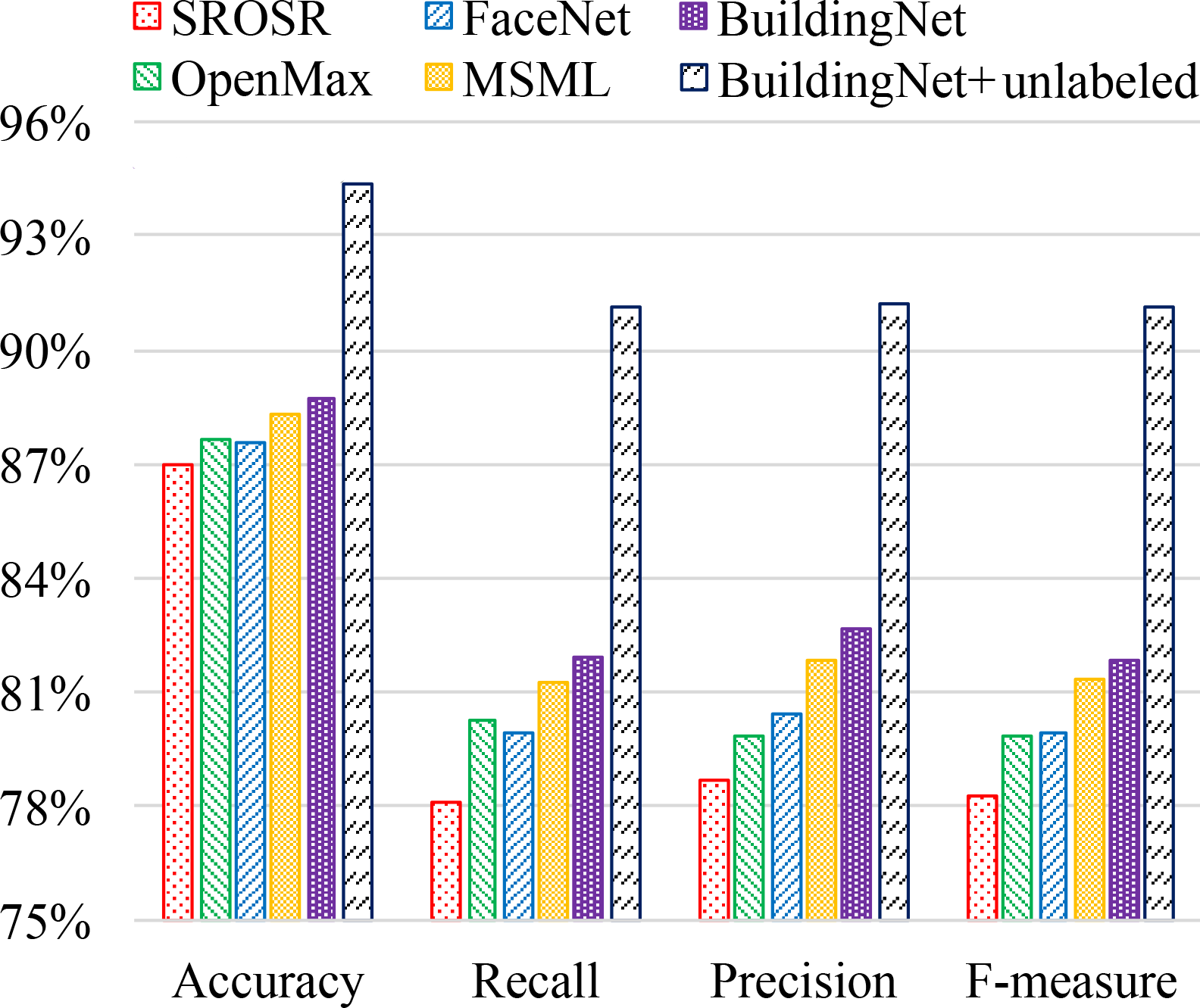}
		\caption{}
	\end{subfigure}\;
	\begin{subfigure}[b]{0.49\linewidth}
		\includegraphics[width=\linewidth]{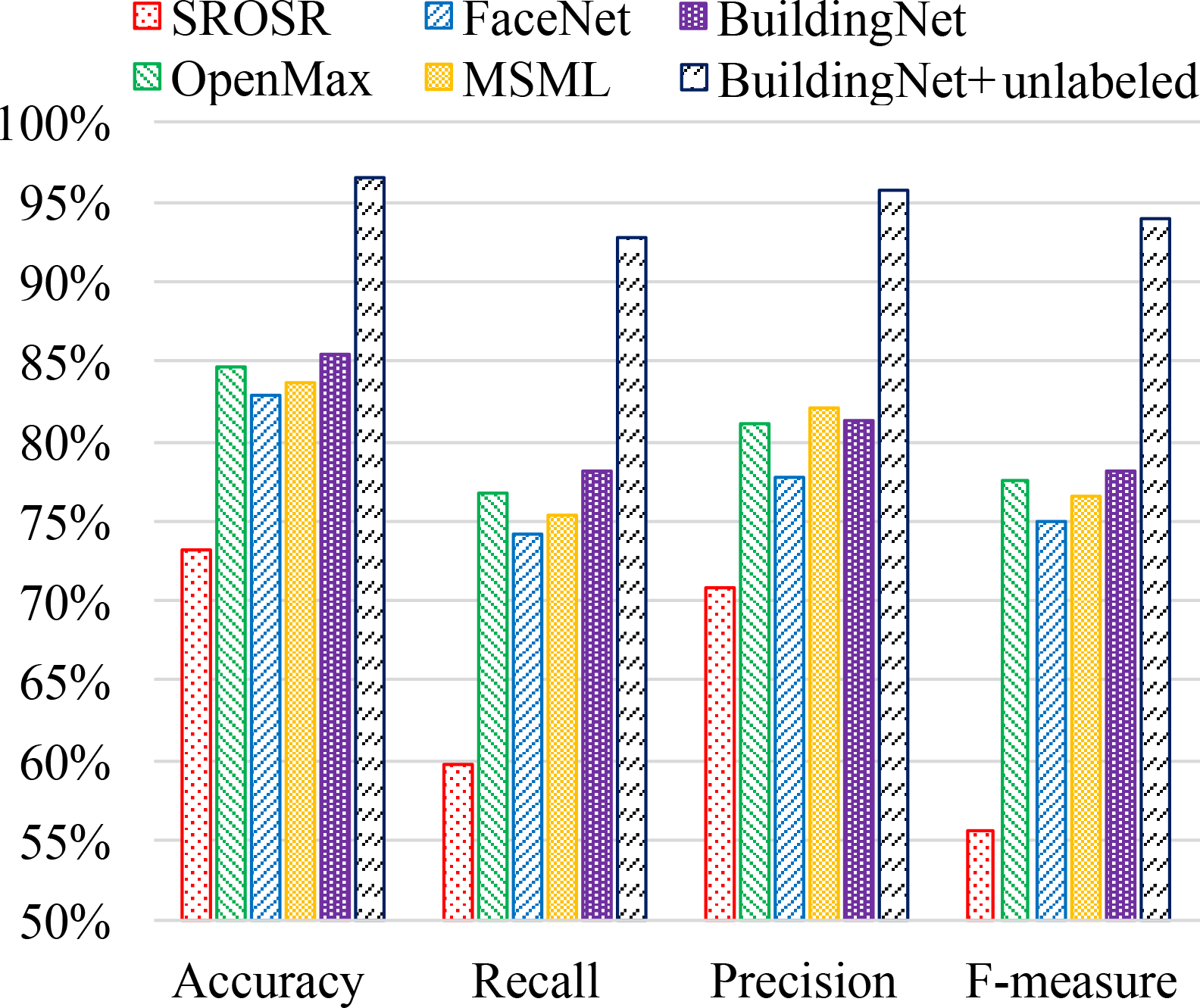}
		\caption{}
	\end{subfigure}
	\caption{Effectiveness of BuildingNet on (a) corner classification and (b) roofline classification (best view in color).}
	\label{fig:rcr}
\end{figure}

\textbf{Hyperparameters.} For the OpenMax, we use the LeNet 5 model to train on the building corner and roofline dataset for 10K iterations with the default setting in Caffe~\cite{jia2014caffe}. We then apply the last fully connected layer to OpenMax for classification.
For our BuildingNet, we pre-train LeNet 5 with the MNIST dataset and fine-tune it with our collected building corner and roofline images. Further, since BuildingNet can also take advantage of unlabeled data (known unknown~\cite{scheirer2014probability}) during training, we also pre-train a LeNet 5 model based on MNIST dataset (0 to 4 as the labeled data and 5 to 9 as the unlabeled data) and fine-tune it with our data. We set the learning rate as 0.1 with the decay rate of 0.95 after each 1K iterations (50K iterations in total). The batch size is 30 images for each class, the embeddings that BuildingNet learns are 128-dimensional, and the $\alpha$ in the triplet relative loss function is 0.5.
We perform 10-fold cross-validation on the models tested, and then compute the accuracy, precision, recall, and F$_1$ score of different models, which are summarized in Figure~\ref{fig:rcr}.


On the corner dataset, BuildingNet achieves an accuracy of 94.34\%, and its recall, precision, and F$_1$ score are all over 91\% when using both labeled and unlabeled data for training. Compared with SROSR and OpenMax, BuildingNet improves the accuracy and F$_1$ score by more than 6\% and 10\%, respectively. When trained with labeled data only, BuildingNet still has the highest accuracy and F$_1$ score (i.e., 88.72\% and 81.8\%), which are 1.1\% and 2\% higher than OpenMax, respectively. Compared with the two loss functions in MSML and FaceNet which are also based on triplet selection, our proposed loss function can improve the accuracy and F$_1$ score by more than 0.4\% and 0.5\%, respectively. For the roofline dataset, the proposed BuildingNet again outperforms the baseline models consistently. These confirm the effectiveness of BuildingNet. 

\begin{figure}[tp]
	\centering
	\setlength{\abovecaptionskip}{1pt}   
	\setlength{\belowcaptionskip}{-2pt}   
	\begin{subfigure}[b]{0.32\linewidth}
		\includegraphics[width=\linewidth]{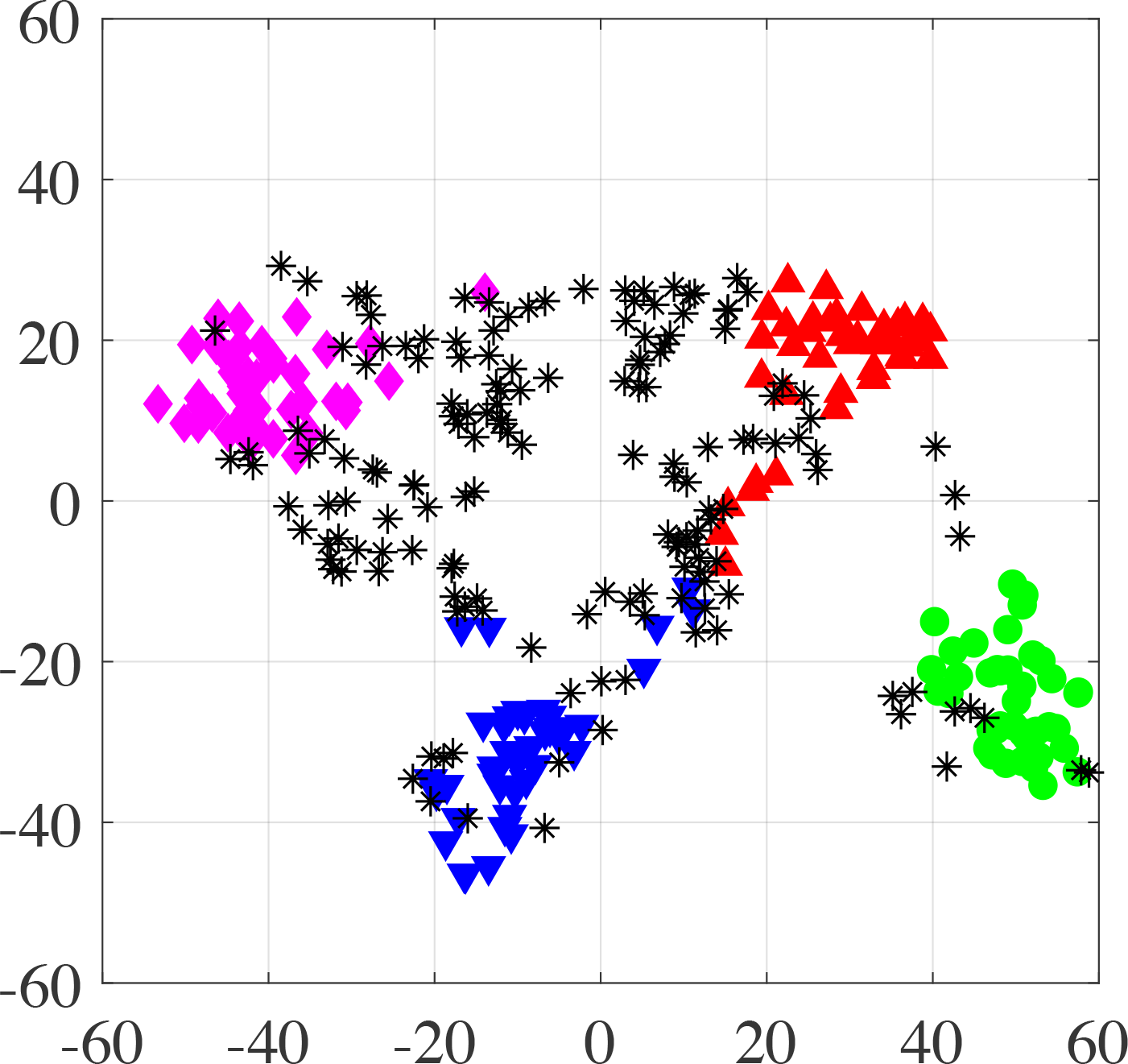}
		\caption{}
	\end{subfigure}
	\begin{subfigure}[b]{0.32\linewidth}
		\includegraphics[width=\linewidth]{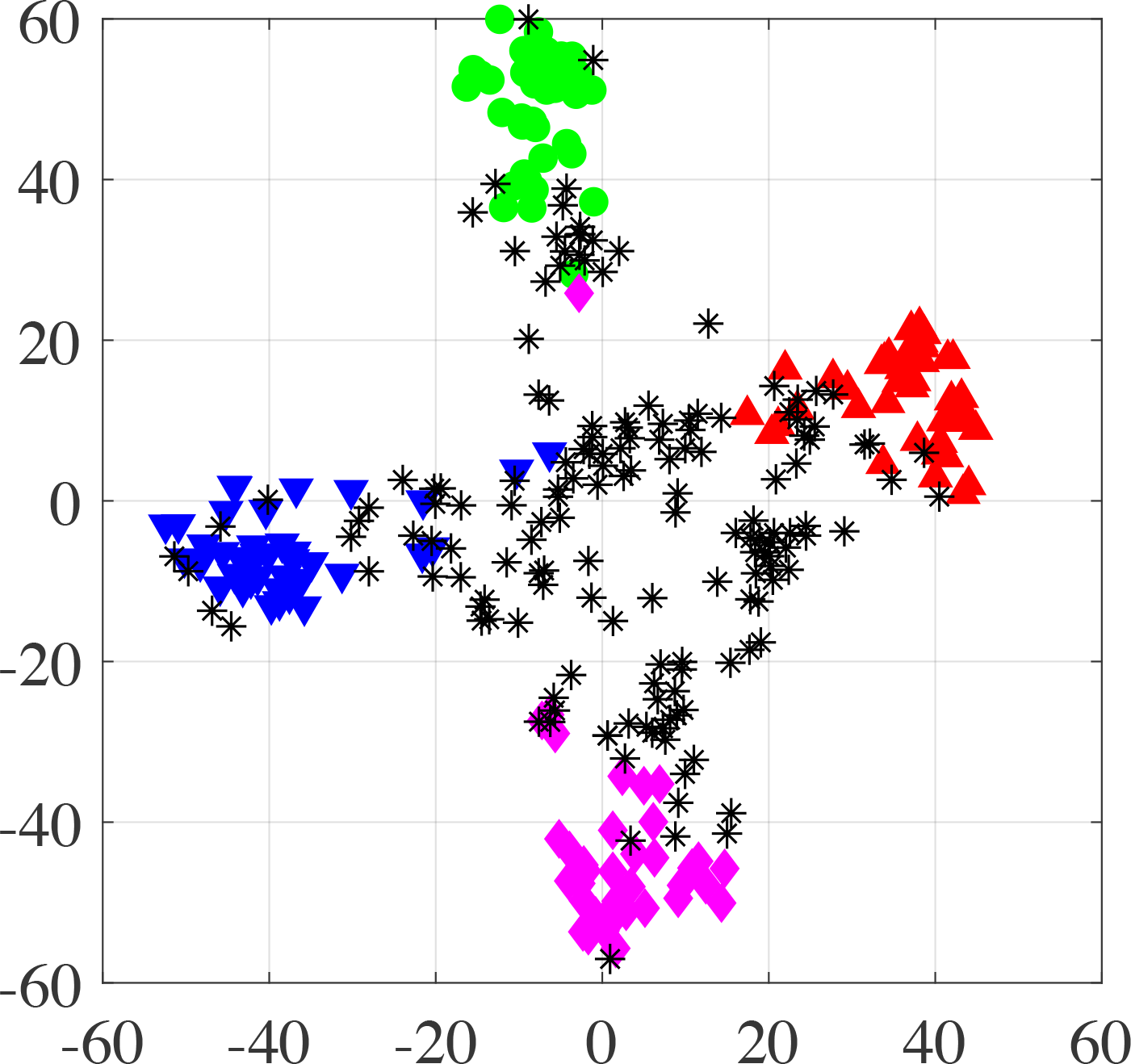}
		\caption{}
	\end{subfigure}
	\begin{subfigure}[b]{0.32\linewidth}
		\includegraphics[width=\linewidth]{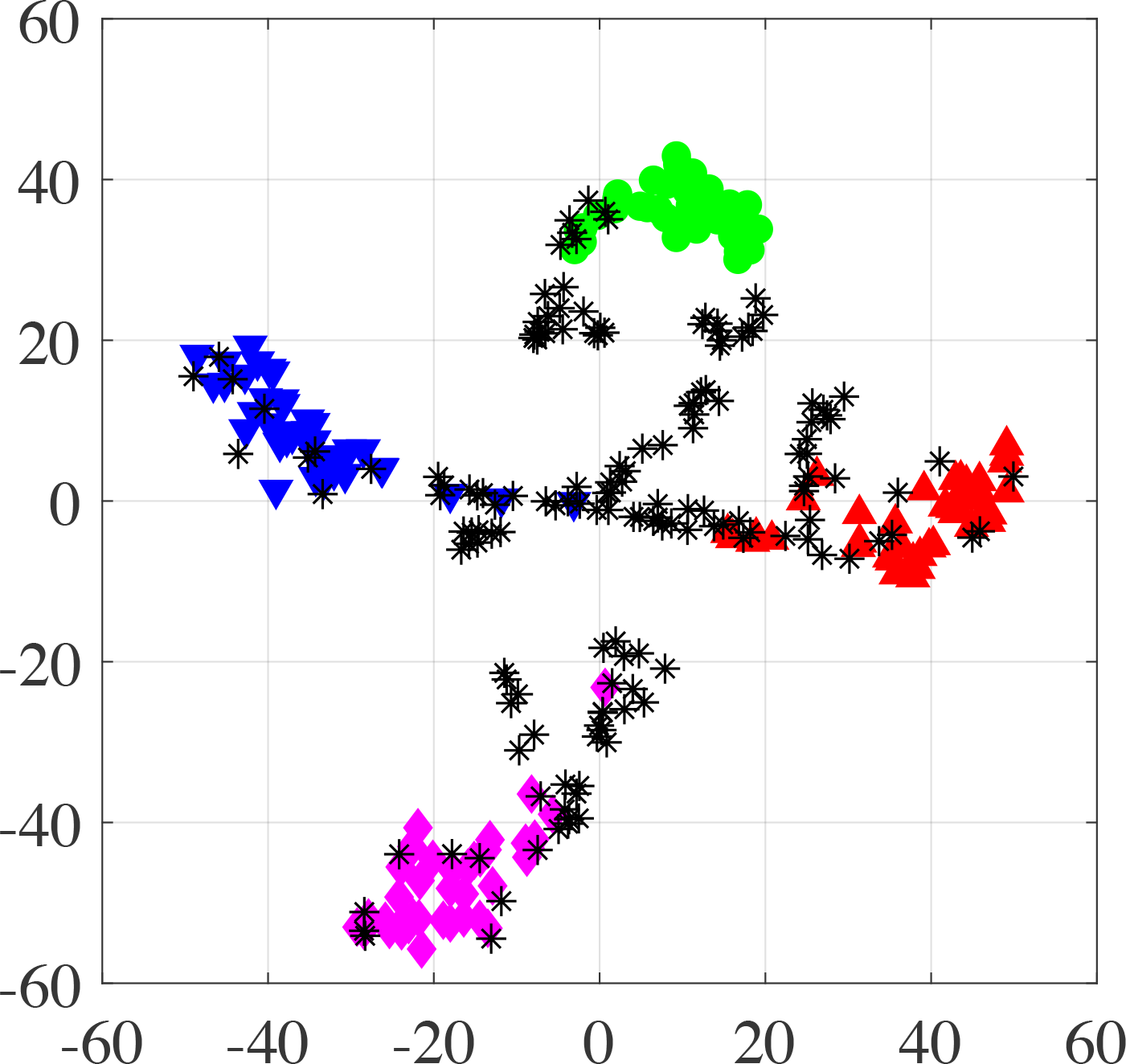}
		\caption{}
	\end{subfigure}
	\caption{t-SNE~\cite{maaten2008visualizing} 2D embeddings of four types of corners and the unlabeled data after 100 epochs, learned by the loss functions in (a) FaceNet, (b) MSML, and (c) the proposed triplet relative loss (best view in color).}
	\label{fig:41}
\end{figure}

To further illustrate the effectiveness of BuildingNet, we visualize the embeddings generated by three triplet based loss functions on the corner dataset, as shown in Fig.~\ref{fig:41}. Compared with random triplet selection with margin (FaceNet) and hard triplet selection with margin (MSML), our triplet relative loss function obtains better classification result with smaller average intra-class distance and larger average inter-class distance after the same number of epochs.

\subsection{Effectiveness of CBHE}
\label{sec:experiment_low}
We evaluate the performance of CBHE on City Blocks and Tall Buildings in this subsection.

\subsubsection{Building height estimation on City Blocks} 

\begin{figure}[b]
	\centering
	\setlength{\abovecaptionskip}{1pt}   
	\setlength{\belowcaptionskip}{-2pt}   
	\begin{subfigure}[b]{0.45\linewidth}
		\includegraphics[width=\linewidth]{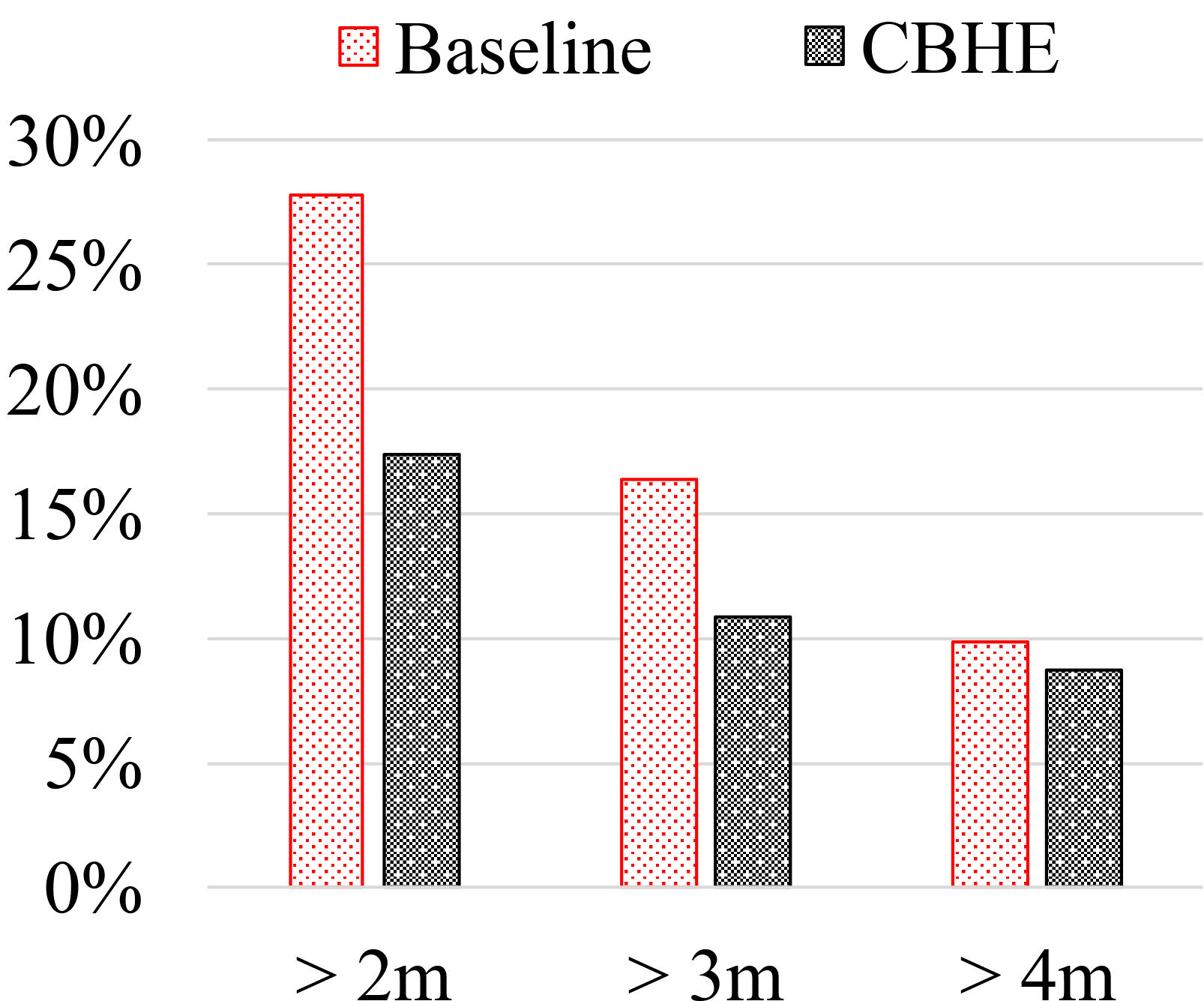}
		\caption{City block 1}
	\end{subfigure}\;\;\;
	\begin{subfigure}[b]{0.45\linewidth}
		\includegraphics[width=\linewidth]{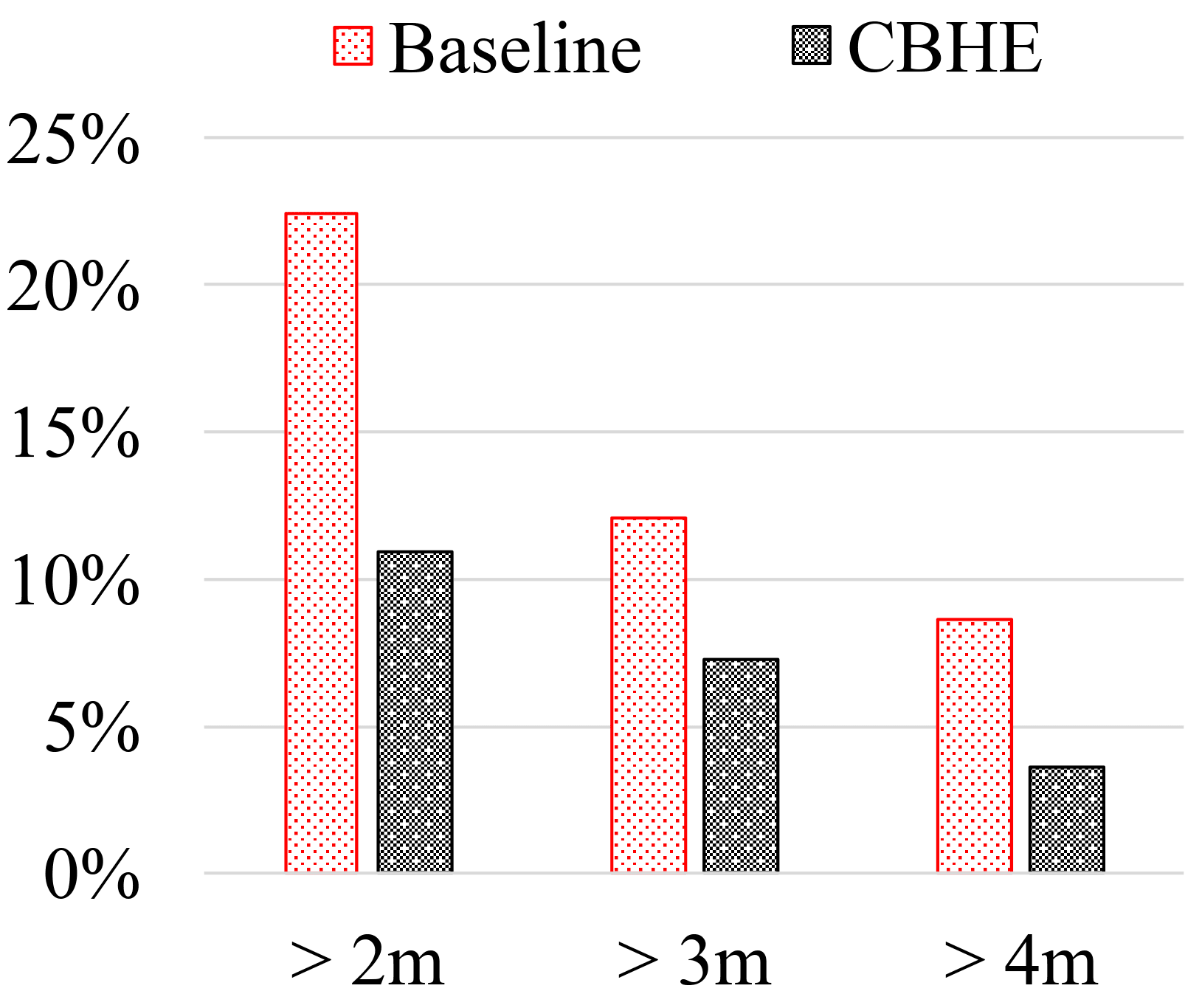}
		\caption{City block 2}
	\end{subfigure}
	\caption{The errors of the baseline method~\cite{Yuan2016} and CBHE on City Blocks.}
	\label{fig:cb}
\end{figure}

Figure~\ref{fig:cb} shows the building high estimation errors of the baseline method [46] and CBHE over the City Blocks dataset. It shows the percentage of buildings where the height estimation is greater than 2, 3, and 4 meters, respectively. In both city blocks, CBHE achieves a smaller percentage of buildings than that of the baseline [46]. 

\begin{figure}[tp]
	\centering
	\setlength{\abovecaptionskip}{1pt}   
	\setlength{\belowcaptionskip}{-2pt}   
	\begin{subfigure}[b]{0.48\linewidth}
		\centering
		\includegraphics[width=0.49\linewidth]{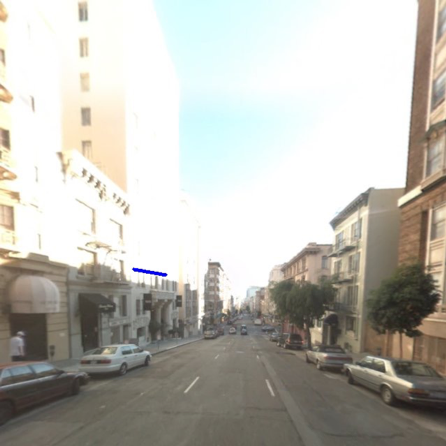}%
		\hfill
		\includegraphics[width=0.49\linewidth]{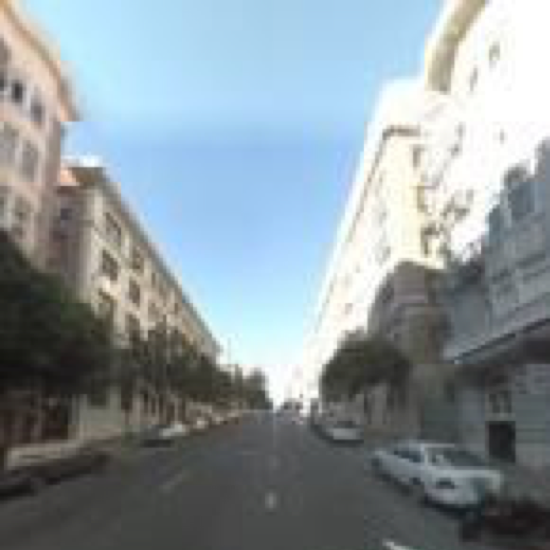}
		\caption{Images used by the baseline }
	\end{subfigure}
	\begin{subfigure}[b]{0.48\linewidth}
		\centering
		\includegraphics[width=0.49\linewidth]{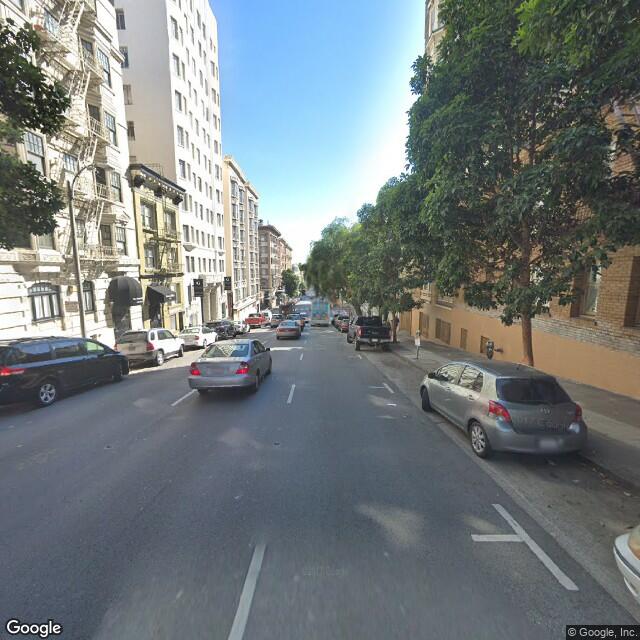}%
		\hfill
		\includegraphics[width=0.49\linewidth]{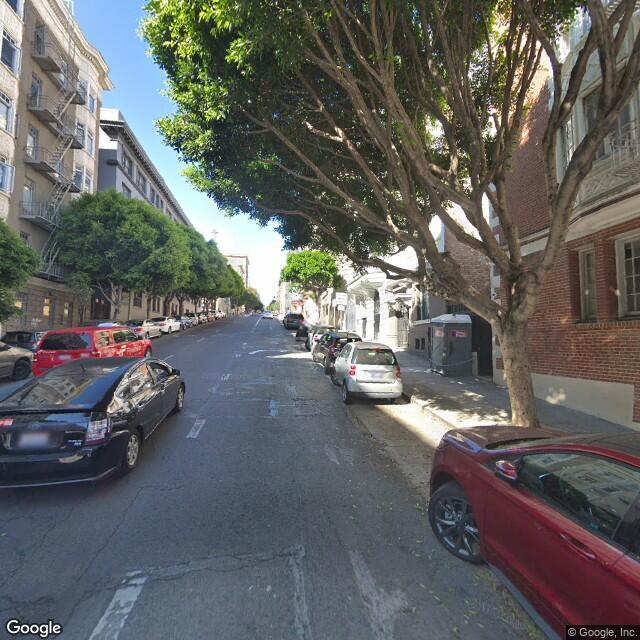}
		\caption{Images used by CBHE}
	\end{subfigure}
	\caption{City block street scene images at the same spots.}
	\label{fig:10}
\end{figure}

In particular, in the first city block (Fig.~\ref{fig:cb}a, which has been used in~\cite{Yuan2016}), CBHE has 10.4\%, 5.5\%, and 1.2\% fewer buildings than those of the baseline with height estimation errors greater than 2, 3, and 4 meters, respectively.  Note that the results of the baseline method are obtained from their paper~\cite{Yuan2016} since we are unable to obtain their source code. Also, even though CBHE is run on the same city block as the baseline in this set of experiments, the images that we used are more challenging to handle as the trees in the street scenes have grown larger which block the buildings (cf. Fig.~\ref{fig:10}). 

Fig~\ref{fig:cb}b shows the result in a second city block (which was not used in~\cite{Yuan2016}). As we are unable to obtain the source code of the baseline method, the result is based on our implementation of their method. CBHE again outperforms the baseline. It has 11.5\%, 4.8\%, and 5\% fewer buildings than those of the baseline with height estimation errors greater than 2, 3, and 4 meters, respectively.  

\subsubsection{Building height estimation on Tall Buildings.}
\label{sec:experiment_tal}
\begin{figure}[h!]
	\centering
	\setlength{\abovecaptionskip}{1pt}   
	\setlength{\belowcaptionskip}{-2pt}   
	\begin{subfigure}[b]{0.24\linewidth}
		\includegraphics[width=\linewidth]{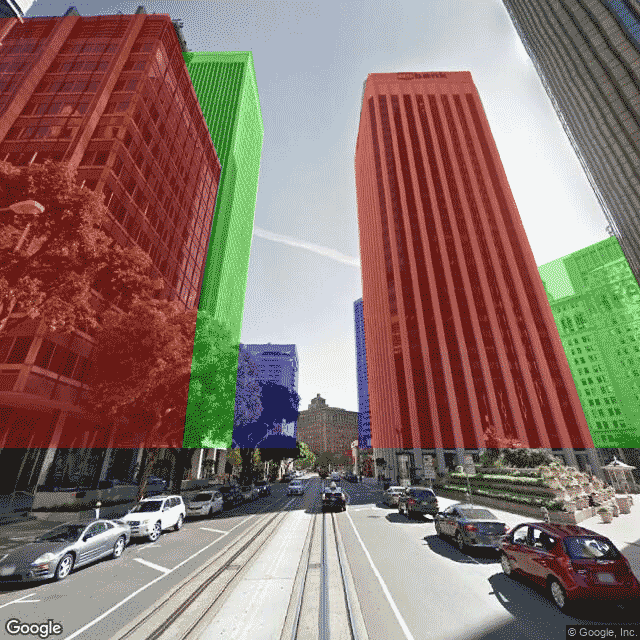}
	\end{subfigure}
	\begin{subfigure}[b]{0.24\linewidth}
		\includegraphics[width=\linewidth]{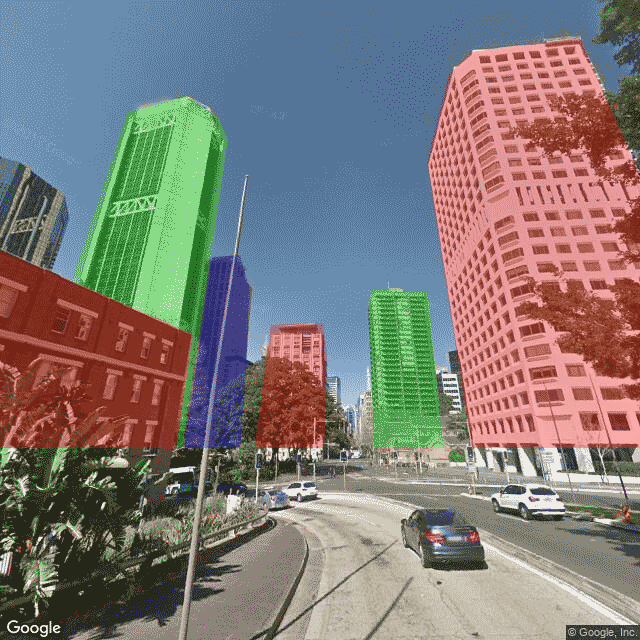}
	\end{subfigure}
	\begin{subfigure}[b]{0.24\linewidth}
		\includegraphics[width=\linewidth]{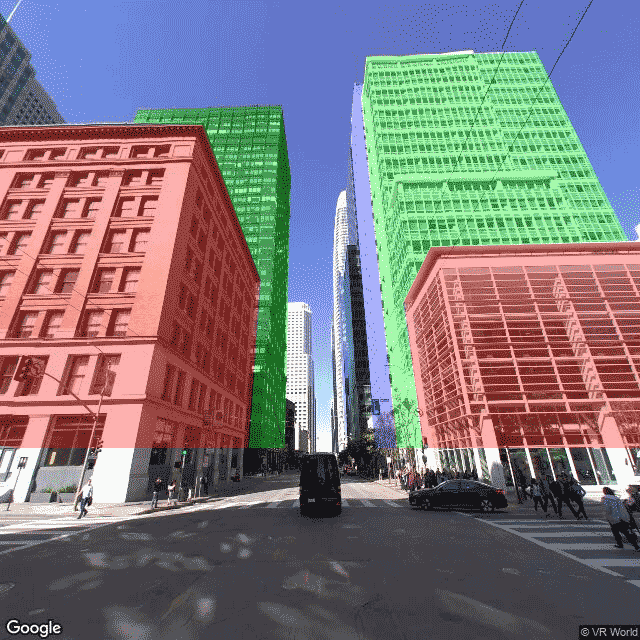}
	\end{subfigure}
	\begin{subfigure}[b]{0.24\linewidth}
		\includegraphics[width=\linewidth]{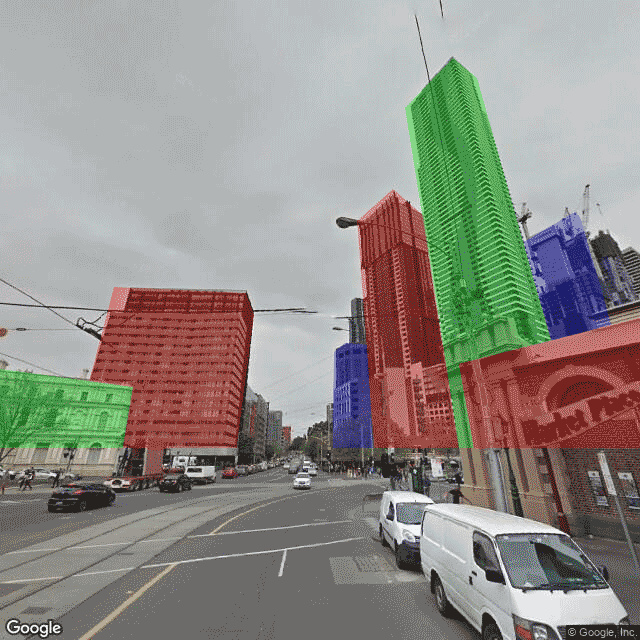}
	\end{subfigure}
	\caption{Tall building examples (best view in color).}
	\label{fig:52}
\end{figure}

For tall buildings, the camera needs to be placed far away with an upward-look view to capture the building roofline. We capture the building images 250 meters away from the buildings via Google Street View API. Fig.~\ref{fig:52} presents examples of the street scene images for tall building height estimation. For each street scene image, we first rotate it to the horizontal view according to Equation~\ref{for:A}, and then compute the height of the buildings according to Section~\ref{sec:heightEst}.

The baseline method~\cite{Yuan2016} cannot be applied to tall buildings, and here we only show the result of CBHE. As shown in Table~\ref{tab:tallBuilding}, more than 53\% of the tall buildings have a height estimation error of less than five meters and 73.33\% of the tall buildings have an error of less than 10 meters. 

\begin{table}[h!]
	\setlength{\abovecaptionskip}{4pt}   
	\caption{CBHE for tall building height estimation.}
	\label{tab:tallBuilding}
	\centering
	\begin{tabular}{cccc}
		\toprule
		Absolute error & Percentage & Relative Error & Percentage \\
		\midrule
		>5m &  45.9\% & >5\% & 40.5\% \\
		>10m &  27.0\% & >10\% & 13.5\% \\
		\bottomrule
	\end{tabular}
\end{table}

The errors of tall buildings may seem larger due to the camera projection (i.e., the errors are multiplied by a larger multiplier for tall buildings). However, we would like to emphasize that the relative errors are still quite low, e.g., since the tall buildings are taller than 100 meters, even a 10-meter error is less than 10\% and is barely notable in reality.  


\subsection{Error Analysis}
We summarize the challenging cases for CBHE in this section. These challenging scenarios will be explored in future work.

\begin{figure}[tp]
	\centering
	\setlength{\abovecaptionskip}{1pt}   
	\setlength{\belowcaptionskip}{-2pt}   
	\begin{subfigure}[b]{0.24\linewidth}
		\includegraphics[width=\linewidth]{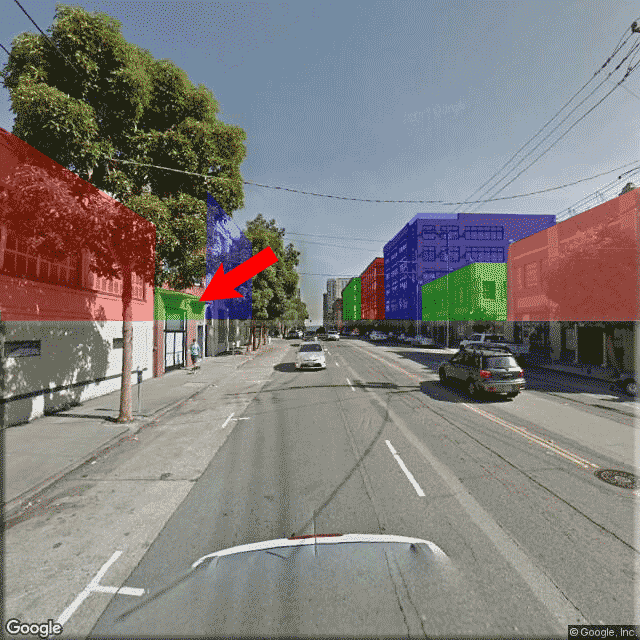}
		\caption{}
	\end{subfigure}
	\begin{subfigure}[b]{0.24\linewidth}
		\includegraphics[width=\linewidth]{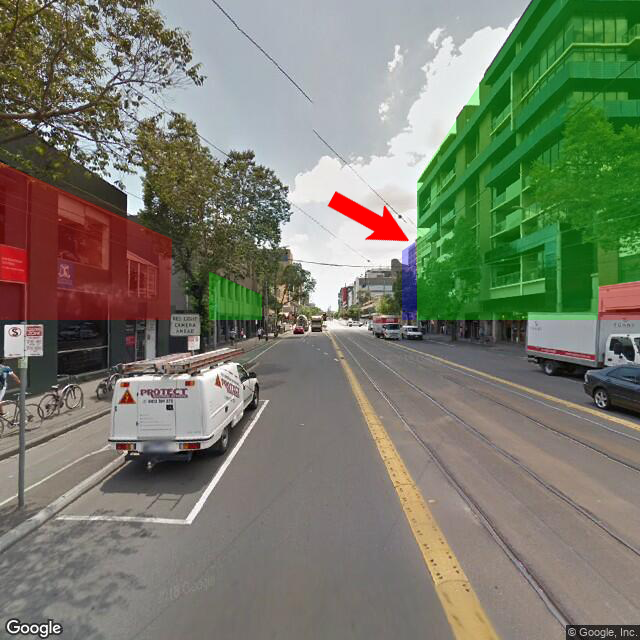}
		\caption{}
	\end{subfigure}
	\begin{subfigure}[b]{0.24\linewidth}
		\includegraphics[width=\linewidth]{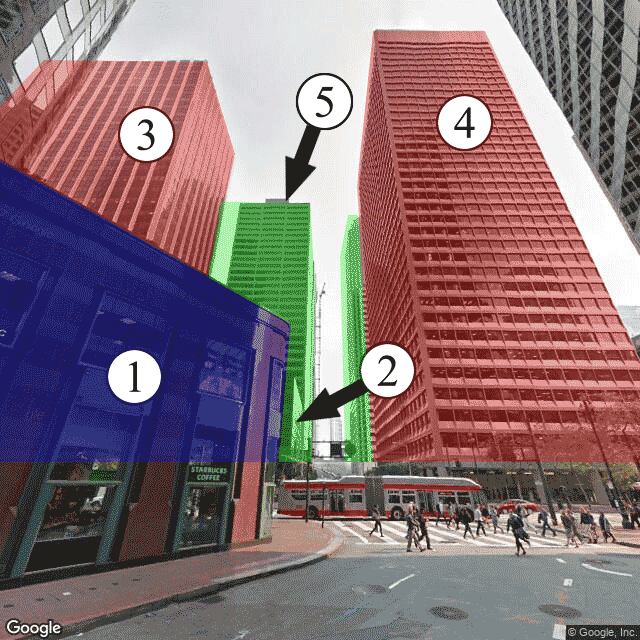}
		\caption{}
	\end{subfigure}
	\begin{subfigure}[b]{0.24\linewidth}
		\includegraphics[width=\linewidth]{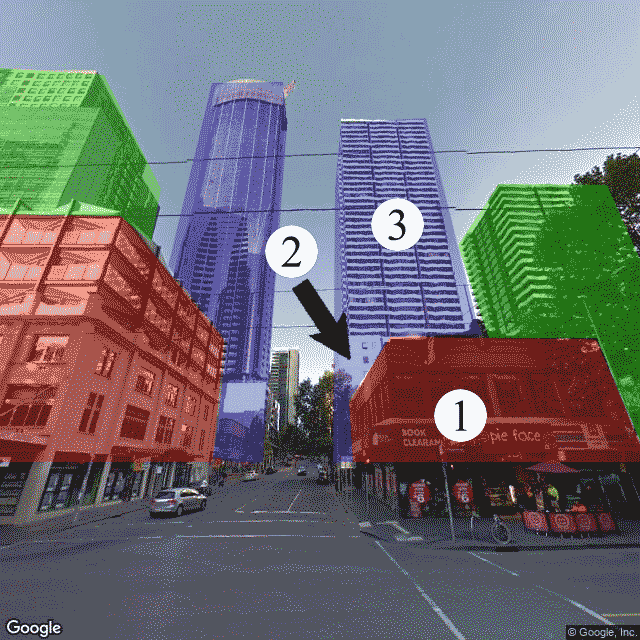}
		\caption{}
	\end{subfigure}
	\caption{Challenging examples (best view in color).}
	\label{fig:63}
\end{figure}

For those buildings whose rooflines are entirely blocked by other objects such as trees, CBHE will ignore them, or output a wrong estimation. Take Fig.~\ref{fig:63}a as an example, the trees on the left-hand side of the image block the roof of the green colored building heavily, resulting in a line below the roof to be identified as the roofline. Additionally, if the corners of a building are not detectable, lines from other buildings behind this building may also impact the result. As illustrated in Fig.~\ref{fig:63}b, the roofline of a building behind the blue colored building was detected as its roofline.

In dense city areas, the buildings may overlap with each other, and it is difficult to match all buildings with their boundaries in a 2D map accurately. Take Fig.~\ref{fig:63}c as an example, building $\textcircled{2}$ is blocked by building $\textcircled{1}$ and building $\textcircled{2}$'s corners have a similar horizontal position to building $\textcircled{5}$. Therefore, CBHE regards the rooflines of building $\textcircled{5}$ as the rooflines of building $\textcircled{2}$, which results in the estimated height of building $\textcircled{2}$ being 77.41m, although its real height is 24.53m. Moreover, the height of building $\textcircled{5}$ is also wrong because the incorrect rooflines of building $\textcircled{2}$ block the rooflines of building $\textcircled{5}$. In Fig.~\ref{fig:63}d, the blue shaded building mask on the right-hand side is wrongly assigned to the building $\textcircled{2}$ (between building $\textcircled{1}$ and building $\textcircled{3}$) because it is closer to the camera with the similar position to building $\textcircled{3}$.

\section{Conclusions}
\label{sec:conclusion}
We proposed a corner-based algorithm named CBHE to learn building height from complex street scene images. CBHE consists of camera location calibration and building roofline detection as its two main steps. To calibrate camera location, CBHE performs camera projection by matching two building corners in street scene images with their physical locations obtained from a 2D map. 
To identify building rooflines, CBHE first detects roofline candidates according to the building footprints in 2D maps and the calibrated camera location. Then, it uses a deep neural network named BuildingNet that we proposed to check whether a roofline candidate indeed is a building roofline. Finally, CBHE ranks the valid rooflines based on an entropy-based ranking algorithm, which also involves building corner information as an essential indicator, and then computes the building height through camera projection. Experimental results show that the proposed BuildingNet model outperforms two state-of-the-art classifiers SROSR and OpenMax consistently, and CBHE outperforms the baseline algorithm by over 10\% in building height estimation accuracy. 

\setlength{\parskip}{-1pt}
\section{acknowledgments}
\label{sec:acknowledgments}
We thank the anonymous reviewers for their feedback. 
We appreciate the valuable discussion with Bayu Distiawan Trsedya, Weihao Chen, and Jungmin Son.
Yunxiang Zhao is supported by the Chinese Scholarship Council (CSC). 
This work is supported by Australian Research Council (ARC) Discovery Project DP180102050, Google Faculty Research Award, and the National Science Foundation of China (Project No. 61402155).
\newpage

\bibliographystyle{ACM-Reference-Format}
\balance 
\bibliography{main}

\end{document}